\pdfoutput=1
\documentclass[10pt,twocolumn,letterpaper]{article}

\usepackage[final]{iccv}      
\usepackage{graphicx}	
\usepackage{amsmath}	
\usepackage{amssymb}	
\usepackage{booktabs}
\usepackage{times}
\usepackage{microtype}
\usepackage{epsfig}
\usepackage{caption}
\usepackage{float}
\usepackage{placeins}
\usepackage{color, colortbl}
\usepackage{stfloats}
\usepackage{enumitem}
\usepackage{tabularx}
\usepackage{xstring}
\usepackage{multirow}
\usepackage{xspace}
\usepackage{url}
\usepackage{subcaption}
\usepackage[hang,flushmargin]{footmisc}
\usepackage{bigstrut} 
\usepackage{makecell}
\usepackage[accsupp]{axessibility}

\definecolor{iccvblue}{rgb}{0.21,0.49,0.74}
\usepackage[pagebackref,breaklinks,colorlinks,allcolors=iccvblue]{hyperref}

\def\paperID{10907} 
\def\confName{ICCV}
\def\confYear{2025}

\title{MP-HSIR: A Multi-Prompt Framework for Universal Hyperspectral Image Restoration}

\author{
Zhehui Wu$^{1}$ \quad
Yong Chen$^{2}$ \quad
Naoto Yokoya$^{3,4}$ \quad
Wei He$^{1,4}$\thanks{Corresponding author}\\  
$^{1}$ Wuhan University \quad
$^{2}$ Jiangxi Normal University \quad
$^{3}$ The University of Tokyo \\
$^{4}$ RIKEN Center for Advanced Intelligence Project \\
{\tt\small \{wuzhehui, weihe1990\}@whu.edu.cn,\quad chen\_yong@jxnu.edu.cn,\quad yokoya@k.u-tokyo.ac.jp}
}

\begin{document}
\maketitle
\begin{abstract}

Hyperspectral images (HSIs) often suffer from diverse and unknown degradations during imaging, leading to severe spectral and spatial distortions. Existing HSI restoration methods typically rely on specific degradation assumptions, limiting their effectiveness in complex scenarios. In this paper, we propose \textbf{MP-HSIR}, a novel multi-prompt framework that effectively integrates spectral, textual, and visual prompts to achieve universal HSI restoration across diverse degradation types and intensities. Specifically, we develop a prompt-guided spatial-spectral transformer, which incorporates spatial self-attention and a prompt-guided dual-branch spectral self-attention. Since degradations affect spectral features differently, we introduce spectral prompts in the local spectral branch to provide universal low-rank spectral patterns as prior knowledge for enhancing spectral reconstruction. Furthermore, the text-visual synergistic prompt fuses high-level semantic representations with fine-grained visual features to encode degradation information, thereby guiding the restoration process. Extensive experiments on 9 HSI restoration tasks, including all-in-one scenarios, generalization tests, and real-world cases, demonstrate that MP-HSIR not only consistently outperforms existing all-in-one methods but also surpasses state-of-the-art task-specific approaches across multiple tasks. The code and models are available at \url{https://github.com/ZhehuiWu/MP-HSIR}.

\end{abstract}
\section{Introduction}
\label{sec:intro}

\begin{figure}[t]
    \centering
    \includegraphics[width=0.48\textwidth]{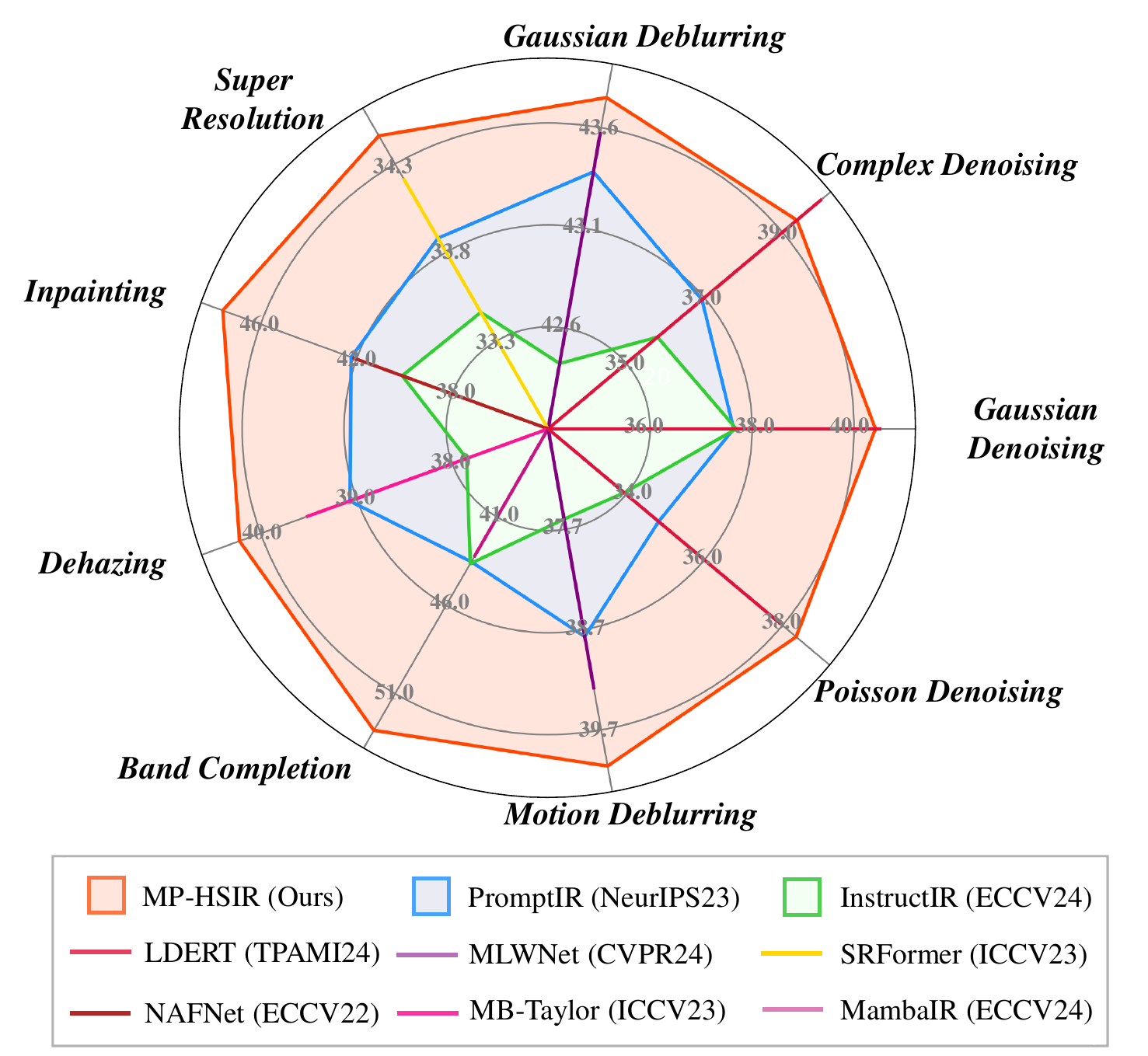} 
    \vskip -10pt
    \caption{PSNR comparison with the state-of-the-art all-in-one and task-specific methods across 9 tasks. Average results across all test datasets. Best viewed in color.}
    \vskip -10pt
    \label{fig:1}
\end{figure}

Hyperspectral images (HSIs) provide significantly higher spectral resolution than RGB images, as illustrated in Figure \ref{fig:2} (a), making them essential for urban planning \cite{heiden2012urban,weber2018hyperspectral}, agricultural production \cite{ravikanth2017extraction,lu2020recent}, and environmental monitoring \cite{andrew2008role,wan2019tailings}. However, due to various environmental factors and imaging limitations, HSIs are often affected by degradations during acquisition, compromising image quality and limiting subsequent analysis. Consequently, HSI restoration is a critical step in the image processing pipeline.

Despite advancements in task-specific HSI restoration methods \cite{he2019non,zhang2023essaformer,wei20203,xu2023aacnet,chen2024hyperspectral,li2024latent},  the diversity and complexity of degradations pose a significant challenge for a single model to handle multiple HSI restoration tasks. As illustrated in Figure \ref{fig:2} (b), different degradation types have distinct effects on the spectral features of HSIs. For instance, noise increases spectral fluctuations, compression reduces spectral reflectance in certain bands, and haze shifts the spectral curve globally. These diverse impacts hinder the generalization capability of task-specific methods across different scenarios. Therefore, a universal framework is required to address the unique distortions caused by different degradations. 

\begin{figure*}[tp]
    \centering
    \includegraphics[width=1.0\textwidth]{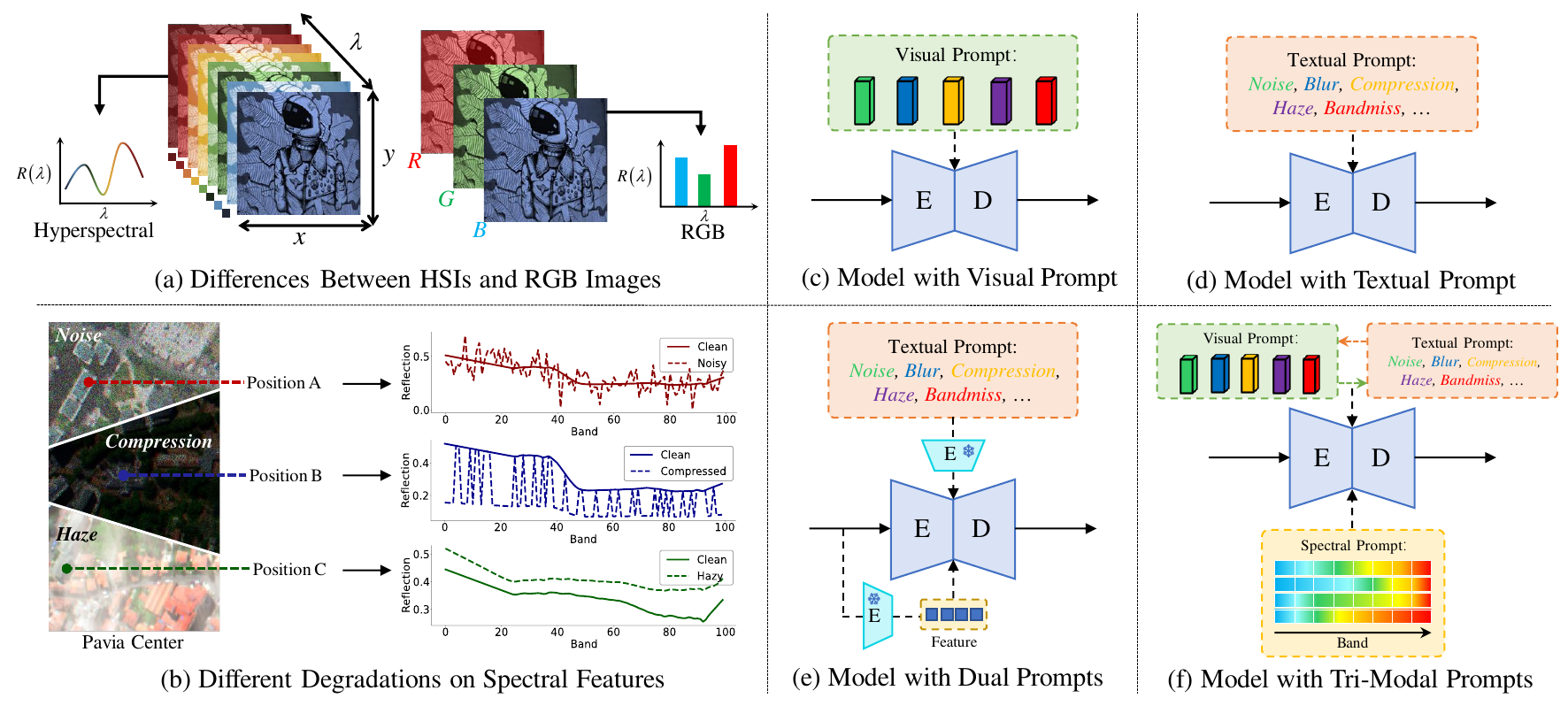} 
    \vskip -10pt
    \caption{Overview of HSI characteristics and prompt-based all-in-one restoration methods. (a) Differences between HSIs and RGB images. (b) Impacts of different degradations on spectral features. (c) Visual prompt-based models. (d) Textual prompt-based models. (e) Dual-modality prompt-based models. (f) The proposed model integrating spectral, textual, and visual prompts. }
    \vskip -5pt
    \label{fig:2}
\end{figure*}

In recent years, all-in-one image restoration methods have achieved
significant progress in RGB images \cite{li2022all,jiang2024survey,wu2024harmony}. Among these, prompt learning-based approaches \cite{potlapalli2023promptir,li2023prompt,conde2024instructir,guo2024onerestore,ai2024multimodal,luo2024photo} have garnered attention for their flexibility in task adaptation and efficient degradation perception. Based on prompt modality, these methods can be categorized into visual, textual, and dual prompts.
As shown in Figure \ref{fig:2} (c), visual prompt-based models \cite{potlapalli2023promptir,li2023prompt} guide restoration by introducing additional prompt components but suffer from limited interpretability. Figure \ref{fig:2} (d) illustrates textual prompt-based models \cite{conde2024instructir,guo2024onerestore}, which convert textual prompts into latent embeddings to assist restoration. Nevertheless, inconsistencies may arise when the semantic gap between text and images is too large. Figure \ref{fig:2} (e) depicts dual prompt-based models \cite{ai2024multimodal,luo2024photo}, which leverage pre-trained vision-language models (VLMs) to extract deep representations from text instructions and degraded images, providing more precise degradation guidance. However, existing VLMs still exhibit limitations in modeling HSI degradation information. Specifically, \textbf{most all-in-one models fail to consider the unique spectral characteristics of HSIs, often resulting in spectral distortions in the restoration process}.

To address the challenges of all-in-one HSI restoration, we propose a universal multi-prompt framework, MP-HSIR, which integrates spectral, textual, and visual prompts, as illustrated in Figure \ref{fig:2} (f). In this framework, spectral prompts guide the spatial-spectral transformer for spectral recovery, while textual and visual prompts synergistically generate degradation-specific information to guide the global restoration. Specifically, the prompt-guided spatial-spectral transformer comprises two core components: spatial self-attention and dual-branch spectral self-attention. The spatial self-attention captures non-local spatial similarities, while the global spectral branch models long-range dependencies in the spectral domain. Meanwhile, the local spectral branch incorporates universal low-rank spectral patterns from the spectral prompts to enhance spectral feature reconstruction. Additionally, the text-visual synergistic prompt deeply fuses the semantic representations of textual prompts with the fine-grained features of visual prompts, enabling controlled restoration while embedding degradation information at the pixel level. As shown in Figure \ref{fig:1}, our method outperforms existing all-in-one methods across 9 HSI restoration tasks and surpasses state-of-the-art task-specific methods in multiple tasks. Overall, our main contributions are as follows:

\begin{itemize}
    \item[$\bullet$] We propose a novel multi-prompt framework for all-in-one HSI restoration, integrating spectral, textual, and visual prompts. The spectral prompt provides universal low-rank spectral patterns to enhance spectral reconstruction.
    \item[$\bullet$] We introduce a text-visual synergistic prompt that combines semantic representations from textual prompts with fine-grained features from visual prompts, enhancing the controllability, interpretability, and degradation adaptability of the HSI restoration process.
    \item[$\bullet$] Extensive experiments on 9 HSI restoration tasks and real-world scenarios demonstrate that MP-HSIR significantly outperforms compared all-in-one methods and surpasses state-of-the-art task-specific approaches in multiple tasks.
\end{itemize}

\section{Related Work}
\label{sec:related}

\subsection{Hyperspectral Image Restoration}

Traditional HSI restoration methods mainly leverage prior knowledge, such as non-local similarity \cite{maggioni2012nonlocal}, low-rank structure \cite{he2019non,liu2021hyperspectral}, and sparsity \cite{xu2022deep,xie2021adaptive}, to constrain the solution space and employ optimization algorithms \cite{shen2021admm,chen2024flex} to solve the inverse problem. Deep learning-based HSI restoration methods directly learn the mapping between degraded and original images through deep neural networks, including convolutional neural networks (CNNs) \cite{yuan2018hyperspectral,wei20203}, transformers \cite{li2023spatial,li2023spectral,yu2023dstrans}, and generative models \cite{li2024latent}. Although these methods have demonstrated strong restoration capabilities, they are typically designed for specific types of degradation, such as noise removal \cite{lai2023hybrid,xiao2024region,bodrito2021trainable}, super-resolution \cite{zhang2023essaformer,wu2023hsr}, and dehazing \cite{ma2022spectral,xu2023aacnet}, lacking universality and adaptability in practical applications.

Recent studies \cite{yu2023dstrans,miao2023dds2m,lai2023hyper,pang2024hir} have explored using a single model for multiple HSI restoration tasks. DDS2M \cite{miao2023dds2m} designs a self-supervised diffusion model, while HIR-Diff \cite{pang2024hir} projects HSI into RGB space to leverage a pre-trained diffusion model for restoration. Nevertheless, these approaches frequently demand intricate hyperparameter tuning and exhibit slow inference speeds. Furthermore, they fail to adequately account for the distinct characteristics of different degradation types, limiting their effectiveness.

\subsection{All-in-One Image Restoration}

All-in-one image restoration \cite{jiang2024survey,wu2024harmony,luo2024photo} aims to establish a unified deep learning framework for restoring images with various degradations. However, developing an effective degradation-aware mechanism to distinguish different degradation remains challenging. AirNet \cite{li2022all} addresses this by employing contrastive learning to extract degradation representations, while HAIR \cite{cao2024hair} leverages a degradation-awareness classifier to capture global information and adaptively generate parameters via a hyper-selection network.

Recently, prompt learning \cite{lu2022prompt,lei2024prompt} has emerged as a promising paradigm for all-in-one image restoration due to its inherent flexibility and task-specific adaptability. PromptIR \cite{potlapalli2023promptir} integrates visual prompts for dynamic guidance, while PIP \cite{li2023prompt} refines them into degradation-aware and basic restoration prompts. To enable more controllable restoration, InstructIR \cite{conde2024instructir} utilizes manually crafted instructions to guide the restoration process. Moreover, with the rise of VLMs, several studies \cite{ai2024multimodal,luo2024photo,qi2024spire} have explored the use of pre-trained VLMs to encode both textual and visual information for identifying degradation types. In the field of all-in-one HSI restoration, PromptHSI \cite{lee2024prompthsi} introduces a text-to-feature modulation mechanism that applies textual prompts to composite degradation restoration in HSI, but it remains inherently limited to a single degradation mode. Overall, existing prompt-based all-in-one methods struggle to effectively characterize different degradation types in HSIs and lack guidance for spectral feature restoration.

\begin{figure*}[tp]
    \centering
    \includegraphics[width=1.0\textwidth]{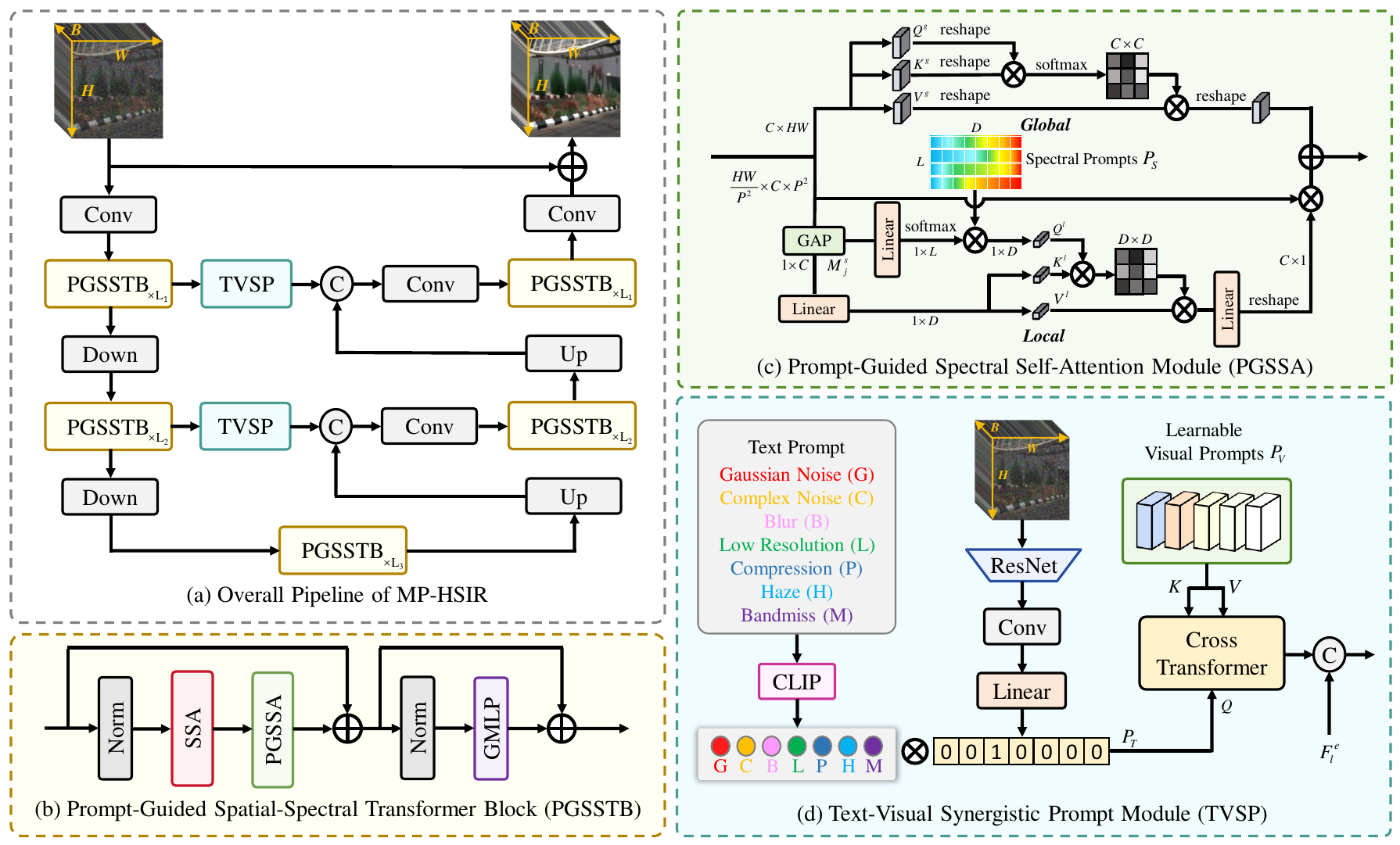} 
    \vskip -5pt
    \caption{(a) The architecture of the proposed MP-HSIR. (b) Prompt-Guided Spatial-Spectral Transformer Block (PGSSTB). (c) Design of the Prompt-Guided Spectral Self-Attention (PGSSA) Module. (d) Text-Visual Synergistic Prompt (TVSP) Module.}
    \label{fig:3}
\end{figure*}

\section{Method}
\label{sec:method}

In this section, we present the multi-prompt framework for universal HSI restoration. First, we review the task of HSI restoration in \ref{sec:method_1}. Then, we describe the overall architecture of the MP-HSIR framework in \ref{sec:method_2}. Finally, we introduce the Spectral Self-Attention with Prompt Guidance module and the Text-Visual Synergistic Prompts module in \ref{sec:method_3} and \ref{sec:method_4}.

\subsection{Preliminary: Hyperspectral Image Restoration}
\label{sec:method_1}

HSI restoration aims to recover the clean image $\mathcal{X}$ from the degraded observation $\mathcal{Y}$ \cite{ghamisi2017advances,zhang2013hyperspectral}, modeled as:
\begin{equation}
\label{eq1}
\mathcal{Y}=\mathcal{H}(\mathcal{X})+\mathcal{N},
\end{equation}

where $\mathcal{H}$ represents multiplicative degradation operations and $\mathcal{N}$ denotes additive degradation operators.  For the denoising task, $\mathcal{H}$  is an identity operation. For the deblurring task, $\mathcal{\mathcal{H}}$  represents the blurring operation. For the inpainting task, $\mathcal{H}$  denotes a binary mask. In practice, the specific degradation in HSI often involves a combination of unknown multiplicative and additive operators. Without prior knowledge of the degradation type, selecting and deploying a specific restoration method requires maintaining multiple specialized models, inevitably leading to high storage costs and complex model management. 

\subsection{Overall Pipeline}
\label{sec:method_2}

Given the degraded observation $\mathcal{Y} \in \mathbb{R}^{H \times W \times B}$, where $H \times W$ represents the spatial resolution and $B$ denotes the number of spectral bands, MP-HSIR first applies a single-layer convolution to extract shallow features. As illustrated in Figure \ref{fig:3} (a), these features are processed through a three-level hierarchical encoder with downsampling operations. This process reduces spatial resolution while increasing channel capacity to obtain multi-scale deep features ${F}_{l}^{e} \in \mathbb{R}^{H \times W \times C}$, where $C$ is the number of channels. Each encoder-decoder pair consists of multiple Prompt-Guided Spatial-Spectral Transformer Blocks (PGSSTBs), with the number of blocks progressively increasing at deeper levels.

To integrate degradation information into the decoding process, we introduce the Text-Visual Synergistic Prompt (TVSP) module at each skip connection between the encoder and decoder. The TVSP module leverages textual and visual prompts to incorporate semantic guidance and fine-grained details, effectively capturing degradation characteristics for restoration. Finally, the restored HSI is generated via a $3 \times 3$ convolution layer followed by an image-level residual connection. The following sections detail the designs of the PGSSTB and TVSP modules.

\subsection{Prompt-Guided Spatial-Spectral Transformer}
\label{sec:method_3}

The PGSSTB is the core component of MP-HSIR, integrating Spatial Self-Attention (SSA), Prompt-Guided Spectral Self-Attention (PGSSA), and a gated multi-layer perceptron (GMLP) \cite{liu2021pay}. To enhance training stability and convergence speed, layer normalization and skip connections \cite{he2016identity} are incorporated, as shown in Figure \ref{fig:3} (b). SSA computes attention along the spatial dimension, capturing the non-local similarity in HSI. To improve computational efficiency while preserving the ability to model long-range dependencies, a sliding window \cite{liu2021swin} of size \(P \times P\) is adopted.  

While SSA captures spatial dependencies, spectral self-attention models long-range correlations across spectral bands, which is crucial for exploiting HSI’s rich spectral information. To compute spectral similarity, the input features are transposed and reshaped into \(F^{in} \in \mathbb{R}^{C \times HW}\) and processed through two complementary branches: global spectral self-attention and prompt-guided local spectral self-attention. The outputs from both branches are then fused and selectively refined using a GMLP. The following sections detail the two branches of PGSSA.

\noindent \textbf{Global Spectral Self-Attention.} 
Global spectral self-attention captures inter-channel dependencies across the entire HSI, as illustrated in the upper part of Figure \ref{fig:3} (c). Specifically, the input feature \( F^{in} \) is first projected into the \textit{query} \( Q^{g} \in \mathbb{R}^{C \times HW} \), \textit{key} \( K^{g} \in \mathbb{R}^{C \times HW} \), and \textit{value} \( V^{g} \in \mathbb{R}^{C \times HW} \) using \( 1 \times 1 \) point-wise convolutions \( W^{P} \), followed by \( 3 \times 3 \) depth-wise convolutions \( W^{D} \):
\begin{equation}
\label{eq2}
\left[{Q}^g, {K}^g, {V}^g\right]={W}^D {W}^P\left[{F}^{in}, {F}^{in}, {F}^{in}\right].
\end{equation}

To capture diverse spectral relationships across different bands, we further divide \( Q^{g} \), \( K^{g} \), and \( V^{g} \) into multiple heads. Within each head, the global spectral attention map \( A^{g} \in \mathbb{R}^{C \times C} \) is computed using the element-wise dot product between \( Q^{g} \) and \( K^{g} \), formulated as:
\begin{equation}
\begin{aligned}
\label{eq3}
&{A}^{g}=\text{Softmax}\left(\frac {{Q}^{g}\cdot{K}^{g}} {\epsilon}\right), \\
&\text{Attention}\left({Q}^{g},{K}^{g},{V}^{g}\right)={W}^{P}\left({A}^{g}{V}^{g}\right),
\end{aligned}
\end{equation}
where $\epsilon$ is a learnable scaling factor that normalizes the results of the dot product.

\noindent \textbf{Prompt-Guided Local Spectral Self-Attention.} Unlike global spectral self-attention, local spectral self-attention models spectral channel dependencies within localized spatial regions, as illustrated in the lower part of Figure \ref{fig:3} (c). Specifically, the input feature \( F^{in} \) is divided into \( \frac{HW}{P^2} \) non-overlapping local patches \( \{ M_1^{in}, M_2^{in}, \dots, M_J^{in} \} \), where each patch \( M_j^{in} \in \mathbb{R}^{C \times P^2} \) has the same spatial size as the spatial self-attention window. To extract representative spectral features \( M_j^{s} \in \mathbb{R}^{1 \times C} \) from each patch, global average pooling is applied along the spatial dimension:
\begin{equation}
\begin{aligned}
\label{eq4}
&M_j^{s}= \left(\text{ AvgPool }\!\left(M_j^{in}\right)\right)^{\top}.
\end{aligned}
\end{equation}

However, due to the susceptibility of spectral structures to various degradations, purely local modeling may lead to an overly dispersed attention distribution, preventing the model from focusing on the most informative spectral channels. To address this, a learnable spectral prompt, \( P_S \in \mathbb{R}^{L \times D} \), is introduced to guide the attention mechanism toward essential spectral patterns, where \( L \) denotes the number of universal low-rank spectral patterns in \( P_S \), and \( D \) represents the dimension. Specifically, \( M_j^{s} \) undergoes a linear transformation followed by Softmax normalization to produce weights. These weights are then used to compute a weighted sum of \( P_S \), which is subsequently projected to obtain the query \( Q^l \). Meanwhile, \( M_j^{s} \) is projected into the same low-dimensional space and further transformed to generate the key \( K^l \) and value \( V^l \), formulated as:
\begin{equation}
\begin{aligned}
\label{eq5}
&Q^{l}= \text{ Softmax }\!\left(M_j^{s} W_{1}^{l} \right) P_{S} W_{3}^{l} , \\
&\left[{K}^l, {V}^l\right]=M_j^{s} W_{2}^{l} W_{3}^{l},
\end{aligned}
\end{equation}
where the linear layer weights \( W_1^l \), \( W_2^l \), and \( W_3^l \) have dimensions \( C \times L \), \( C \times D \), and \( D \times D \), respectively.

The spectral prompt $P_S$, learned during training, provides universally applicable low-rank spectral patterns for local spectral self-attention, enabling any local spectral feature to be represented as a linear combination of these patterns. This enhances the performance of local spectral modeling and facilitates spectral information recovery, thereby improving the model’s generalization across different datasets and degradation scenarios. Finally, the local spectral attention map ${A}^{l}\in \mathbb{R}^{D \times D}$ can be obtained by:
\begin{equation}
\begin{aligned}
\label{eq6}
&{A}^{l}=\text{Softmax}\left(\frac {{Q}^{l}\cdot{K}^{l}} {\epsilon}\right), \\
&\text{Attention}\left({Q}^{l},{K}^{l},{V}^{l}\right)={W}^{P}\left({A}^{l}{V}^{l}\right),
\end{aligned}
\end{equation}
where $W^{P}$ remaps the low-dimensional spectral features to the original dimension $M_j^{out} \in \mathbb{R}^{C \times 1}$, which is then used to weight $M_j^{in}$ to obtain the final output.

\subsection{Text-Visual Synergistic Prompt}
\label{sec:method_4}

The spectral prompt exhibits strong robustness to HSIs with varying degradations but lacks explicit guidance on the degradation type. To address this limitation, we introduce the TVSP module, as illustrated in Figure \ref{fig:3} (d), which integrates textual prompts and learnable visual prompts to provide degradation-specific information for the restoration network.

First, we employ a degradation predictor \( \Phi \) \cite{chi2020fast} to classify the degradation type of the input observation (details are provided in Section A of the supplementary material). Based on predefined textual descriptions \( T_{text} \) corresponding to different degradation types, a frozen CLIP model \cite{radford2021learning} generates embeddings that serve as the textual prompt \( P_T \). To bridge the domain gap between RGB and HSI, these embeddings are transformed into learnable parameters, enabling the network to adapt and refine the textual prompts during training for improved alignment with HSI characteristics. The final textual prompt is selected based on the classification result:

\begin{equation}
\label{eq7}
P_T=\Phi(\mathcal{X}) \cdot \operatorname{Clip}\left(T_{text }\right).
\end{equation}

However, the textual prompt $P_T$ provides only global degradation information and lacks pixel-level precision. To address this, we introduce a learnable visual prompt $P_V$ and fuse it with the textual prompt via a cross-attention. This allows $P_T$ to regulate global degradation while $P_V$ refines local features, enabling degradation information to be effectively integrated and adjusted, formally expressed as:
\begin{equation}
\label{eq8}
F_l^{out}=\operatorname{Concat}\left(F_l^{e}, \text{Attention }\!\left(P_T, P_V\right)\right),
\end{equation}
where the output \( F_l^{out} \) is finally passed to the decoder via a skip connection. Compared to VLM-based dual prompts, TVSP is more flexible and adaptive, enabling prompt adjustments for different HSI domains.

\section{Experiments}
\label{sec:experiments}


\begin{table*}[htbp]
  \centering
  \renewcommand{\arraystretch}{1.2} 
  \resizebox{\textwidth}{!}{
    \begin{tabular}{rrrrrrrrrr}
    \toprule
    \addlinespace[0.2em] 
    \hline
    \multicolumn{1}{c|}{\multirow{3}{*}{\textbf{Type}}} & \multicolumn{1}{c|}{\multirow{3}{*}{\textbf{Method}}} & \multicolumn{3}{c|}{\textbf{\textit{Gaussian Denoising (Sigma = 30, 50, 70)}}} & \multicolumn{1}{c|}{\multirow{3}{*}{\textbf{Method}}} & \multicolumn{3}{c}{\textbf{\textit{Complex Denoising (Case = 1, 2, 3, 4)}}}  \\
\cline{3-5}\cline{7-9}    \multicolumn{1}{c|}{} & \multicolumn{1}{c|}{} & \multicolumn{1}{c|}{ICVL \cite{arad2016sparse}} & \multicolumn{1}{c|}{ARAD \cite{arad2022ntire}} & \multicolumn{1}{c|}{Xiong’an \cite{yi2020aerial}} & \multicolumn{1}{c|}{} & \multicolumn{1}{c|}{ICVL \cite{arad2016sparse}} & \multicolumn{1}{c|}{ARAD \cite{arad2022ntire}} & \multicolumn{1}{c}{WDC \cite{zhu2017hyperspectral}}  \\
\cline{3-5}\cline{7-9}    \multicolumn{1}{c|}{} & \multicolumn{1}{c|}{} & \multicolumn{1}{c|}{PSNR / SSIM $\uparrow$} & \multicolumn{1}{c|}{PSNR / SSIM $\uparrow$} & \multicolumn{1}{c|}{PSNR / SSIM $\uparrow$} & \multicolumn{1}{c|}{} & \multicolumn{1}{c|}{PSNR / SSIM $\uparrow$} & \multicolumn{1}{c|}{PSNR / SSIM $\uparrow$} & \multicolumn{1}{c}{PSNR / SSIM $\uparrow$}  \\
    \hline
    \multicolumn{1}{c|}{\multirow{4}[2]{*}{Task
Specific}} & \multicolumn{1}{c|}{QRNN3D \cite{wei20203}} & \multicolumn{1}{c|}{39.99 / 0.947} & \multicolumn{1}{c|}{39.18 / 0.932} & \multicolumn{1}{c|}{36.06 / 0.829} & \multicolumn{1}{c|}{QRNN3D \cite{wei20203}} & \multicolumn{1}{c|}{41.60 / 0.966} & \multicolumn{1}{c|}{41.11 / 0.960} & \multicolumn{1}{c}{30.82 / 0.868}  \\
    \multicolumn{1}{c|}{} & \multicolumn{1}{c|}{SST \cite{li2023spatial}} & \multicolumn{1}{c|}{41.31 / 0.959} & \multicolumn{1}{c|}{40.86 / 0.955} & \multicolumn{1}{c|}{37.53 / 0.850} & \multicolumn{1}{c|}{SST \cite{li2023spatial}} & \multicolumn{1}{c|}{42.43 / 0.971} & \multicolumn{1}{c|}{41.95 / 0.967} & \multicolumn{1}{c}{32.71 / 0.889} \\
    \multicolumn{1}{c|}{} & \multicolumn{1}{c|}{SERT \cite{li2023spectral}} & \multicolumn{1}{c|}{41.55 / \textcolor{blue}{0.967}} & \multicolumn{1}{c|}{\textcolor{blue}{41.09} / 0.959} & \multicolumn{1}{c|}{37.83 / 0.859} & \multicolumn{1}{c|}{SERT \cite{li2023spectral}} & \multicolumn{1}{c|}{\textcolor{blue}{43.31} / \textcolor{blue}{0.976}} & \multicolumn{1}{c|}{\textcolor{blue}{42.87} / \textcolor{blue}{0.973}} & \multicolumn{1}{c}{33.31 / 0.903}  \\
    \multicolumn{1}{c|}{} & \multicolumn{1}{c|}{LDERT \cite{li2024latent}} & \multicolumn{1}{c|}{\textcolor{red}{41.92} / \textcolor{red}{0.969}} & \multicolumn{1}{c|}{\textcolor{red}{41.47} / \textcolor{red}{0.965}} & \multicolumn{1}{c|}{\textcolor{blue}{38.14} / 0.865} & \multicolumn{1}{c|}{LDERT \cite{li2024latent}} & \multicolumn{1}{c|}{\textcolor{red}{43.42} / \textcolor{red}{0.977}} & \multicolumn{1}{c|}{\textcolor{red}{43.02} / \textcolor{red}{0.974}} & \multicolumn{1}{c}{33.49 / 0.904}  \\
    \hline
    \multicolumn{1}{c|}{\multirow{7}[2]{*}{All in One}} & \multicolumn{1}{c|}{AirNet \cite{li2022all}} & \multicolumn{1}{c|}{39.76 / 0.943} & \multicolumn{1}{c|}{39.19 / 0.920} & \multicolumn{1}{c|}{31.94 / 0.668} & \multicolumn{1}{c|}{AirNet \cite{li2022all}} & \multicolumn{1}{c|}{40.68 / 0.959} & \multicolumn{1}{c|}{40.09 / 0.945} & \multicolumn{1}{c}{28.07 / 0.726} 
\\
    \multicolumn{1}{c|}{} & \multicolumn{1}{c|}{PromptIR \cite{potlapalli2023promptir}} & \multicolumn{1}{c|}{40.25 / 0.953} & \multicolumn{1}{c|}{39.69 / 0.945} & \multicolumn{1}{c|}{32.99 / 0.684} & \multicolumn{1}{c|}{PromptIR \cite{potlapalli2023promptir}} & \multicolumn{1}{c|}{41.29 / 0.965} & \multicolumn{1}{c|}{40.71 / 0.954} & \multicolumn{1}{c}{28.83 / 0.735} \\
    \multicolumn{1}{c|}{} & \multicolumn{1}{c|}{PIP \cite{li2023prompt}} & \multicolumn{1}{c|}{40.87 / 0.958} & \multicolumn{1}{c|}{40.25 / 0.952} & \multicolumn{1}{c|}{32.64 / 0.674} & \multicolumn{1}{c|}{PIP \cite{li2023prompt}} & \multicolumn{1}{c|}{41.67 / 0.968} & \multicolumn{1}{c|}{41.17 / 0.962} & \multicolumn{1}{c}{28.50 / 0.724} \\
    \multicolumn{1}{c|}{} & \multicolumn{1}{c|}{HAIR \cite{cao2024hair}} & \multicolumn{1}{c|}{40.51 / 0.956} & \multicolumn{1}{c|}{39.91 / 0.949} & \multicolumn{1}{c|}{32.54 / 0.679} & \multicolumn{1}{c|}{HAIR \cite{cao2024hair}} & \multicolumn{1}{c|}{40.63 / 0.958} & \multicolumn{1}{c|}{40.15 / 0.947} & \multicolumn{1}{c}{28.18 / 0.729} \\
    \multicolumn{1}{c|}{} & \multicolumn{1}{c|}{InstructIR \cite{conde2024instructir}} & \multicolumn{1}{c|}{41.02 / 0.960} & \multicolumn{1}{c|}{40.32 / 0.954} & \multicolumn{1}{c|}{31.74 / 0.666} & \multicolumn{1}{c|}{InstructIR \cite{conde2024instructir}} & \multicolumn{1}{c|}{40.12 / 0.957} & \multicolumn{1}{c|}{39.81 / 0.952} & \multicolumn{1}{c}{27.49 / 0.707} \\
    \multicolumn{1}{c|}{} & \multicolumn{1}{c|}{PromptHSI \cite{lee2024prompthsi}} & \multicolumn{1}{c|}{40.65 / 0.960} & \multicolumn{1}{c|}{40.04 / 0.956} & \multicolumn{1}{c|}{38.07 / \textcolor{blue}{0.881}} & \multicolumn{1}{c|}{PromptHSI \cite{lee2024prompthsi}} & \multicolumn{1}{c|}{39.14 / 0.955} & \multicolumn{1}{c|}{38.75 / 0.936} & \multicolumn{1}{c}{\textcolor{blue}{33.77} / \textcolor{blue}{0.912}} \\
    \multicolumn{1}{c|}{} & \multicolumn{1}{c|}{  MP-HSIR (Ours)} & \multicolumn{1}{c|}{\textcolor{blue}{41.62} / 0.964} & \multicolumn{1}{c|}{\textcolor{blue}{41.09} / \textcolor{blue}{0.960}} & \multicolumn{1}{c|}{\textcolor{red}{38.81} / \textcolor{red}{0.897}} & \multicolumn{1}{c|}{  MP-HSIR (Ours)} & \multicolumn{1}{c|}{42.29 / 0.971} & \multicolumn{1}{c|}{41.99 / 0.969} & \multicolumn{1}{c}{\textcolor{red}{34.07} / \textcolor{red}{0.918}} \\
    \hline
    \addlinespace[0.2em] 
    \hline
    \multicolumn{1}{c|}{\multirow{3}{*}{\textbf{Type}}} & \multicolumn{1}{c|}{\multirow{3}{*}{\textbf{Method}}} & \multicolumn{3}{c|}{\textbf{\textit{Gaussian Deblurring (Radius = 9, 15, 21 \& 7, 11, 15)}}} & \multicolumn{1}{c|}{\multirow{3}{*}{\textbf{Method}}} & \multicolumn{3}{c}{\textbf{\textit{Super-Resolution  (Scale = 2, 4, 8)}}}  \\
\cline{3-5}\cline{7-9}    \multicolumn{1}{c|}{} & \multicolumn{1}{c|}{} & \multicolumn{1}{c|}{ICVL \cite{arad2016sparse}} & \multicolumn{1}{c|}{PaviaC \cite{huang2009comparative}} & \multicolumn{1}{c|}{Eagle \cite{peerbhay2013commercial}} & \multicolumn{1}{c|}{} & \multicolumn{1}{c|}{ARAD \cite{arad2022ntire}} & \multicolumn{1}{c|}{PaviaU \cite{huang2009comparative}} & \multicolumn{1}{c}{Houston \cite{wu2017convolutional}}  \\
\cline{3-5}\cline{7-9}    \multicolumn{1}{c|}{} & \multicolumn{1}{c|}{} & \multicolumn{1}{c|}{PSNR / SSIM $\uparrow$} & \multicolumn{1}{c|}{PSNR / SSIM $\uparrow$} & \multicolumn{1}{c|}{PSNR / SSIM $\uparrow$} & \multicolumn{1}{c|}{} & \multicolumn{1}{c|}{PSNR / SSIM $\uparrow$} & \multicolumn{1}{c|}{PSNR / SSIM $\uparrow$} & \multicolumn{1}{c}{PSNR / SSIM $\uparrow$}  \\
    \hline
    \multicolumn{1}{c|}{\multirow{4}[2]{*}{Task
Specific}} & \multicolumn{1}{c|}{Stripformer \cite{tsai2022stripformer}} & \multicolumn{1}{c|}{46.03 / 0.988} & \multicolumn{1}{c|}{37.13 / 0.913} & \multicolumn{1}{c|}{41.96 / 0.958} & \multicolumn{1}{c|}{SNLSR \cite{hu2024exploring}} & \multicolumn{1}{c|}{36.05 / 0.898} & \multicolumn{1}{c|}{30.55 / 0.730} & \multicolumn{1}{c}{31.70 / 0.787}   \\
    \multicolumn{1}{c|}{} & \multicolumn{1}{c|}{FFTformer \cite{kong2023efficient}} & \multicolumn{1}{c|}{46.65 / 0.988} & \multicolumn{1}{c|}{37.96 / 0.921} & \multicolumn{1}{c|}{42.76 / 0.962} & \multicolumn{1}{c|}{MAN \cite{wang2024multi}} & \multicolumn{1}{c|}{36.88 / 0.911} & \multicolumn{1}{c|}{30.92 / 0.733} & \multicolumn{1}{c}{32.03 / 0.791} \\
    \multicolumn{1}{c|}{} & \multicolumn{1}{c|}{LoFormer \cite{mao2024loformer}} & \multicolumn{1}{c|}{47.15 / \textcolor{blue}{0.989}} & \multicolumn{1}{c|}{37.72 / 0.917} & \multicolumn{1}{c|}{42.59 / 0.962} & \multicolumn{1}{c|}{ESSAformer \cite{zhang2023essaformer}} & \multicolumn{1}{c|}{37.40 / 0.918} & \multicolumn{1}{c|}{31.34 / 0.738} & \multicolumn{1}{c}{32.37 / 0.793} \\
    \multicolumn{1}{c|}{} & \multicolumn{1}{c|}{MLWNet \cite{gao2024efficient}} & \multicolumn{1}{c|}{47.66 / \textcolor{red}{0.990}} & \multicolumn{1}{c|}{\textcolor{red}{39.01} / \textcolor{red}{0.928}} & \multicolumn{1}{c|}{\textcolor{red}{44.04} / \textcolor{red}{0.969}} & \multicolumn{1}{c|}{SRFormer \cite{zhou2023srformer}} & \multicolumn{1}{c|}{38.02 / 0.922} & \multicolumn{1}{c|}{\textcolor{blue}{31.80} / \textcolor{blue}{0.751}} & \multicolumn{1}{c}{32.80 / 0.803}   \\
    \hline
    \multicolumn{1}{c|}{\multirow{7}[2]{*}{All
in
One}} & \multicolumn{1}{c|}{AirNet \cite{li2022all}} & \multicolumn{1}{c|}{47.21 / \textcolor{blue}{0.989}} & \multicolumn{1}{c|}{37.76 / 0.918} & \multicolumn{1}{c|}{42.42 / 0.963} & \multicolumn{1}{c|}{AirNet \cite{li2022all}} & \multicolumn{1}{c|}{36.73 / 0.910} & \multicolumn{1}{c|}{30.93 / 0.735} & \multicolumn{1}{c}{32.00 / 0.791}   \\
    \multicolumn{1}{c|}{} & \multicolumn{1}{c|}{PromptIR \cite{potlapalli2023promptir}} & \multicolumn{1}{c|}{\textcolor{blue}{47.67} / \textcolor{red}{0.990}} & \multicolumn{1}{c|}{38.72 / \textcolor{blue}{0.927}} & \multicolumn{1}{c|}{43.74 / \textcolor{blue}{0.968}} & \multicolumn{1}{c|}{PromptIR \cite{potlapalli2023promptir}} & \multicolumn{1}{c|}{37.37 / 0.918} & \multicolumn{1}{c|}{31.53 / 0.746} & \multicolumn{1}{c}{32.73 / 0.799} \\
    \multicolumn{1}{c|}{} & \multicolumn{1}{c|}{PIP \cite{li2023prompt}} & \multicolumn{1}{c|}{47.52 / \textcolor{red}{0.990}} & \multicolumn{1}{c|}{38.52 / 0.924} & \multicolumn{1}{c|}{43.17 / 0.967} & \multicolumn{1}{c|}{PIP \cite{li2023prompt}} & \multicolumn{1}{c|}{\textcolor{red}{38.36} / \textcolor{red}{0.926}} & \multicolumn{1}{c|}{31.77 / 0.749} & \multicolumn{1}{c}{\textcolor{blue}{32.89} / \textcolor{blue}{0.805}} \\
    \multicolumn{1}{c|}{} & \multicolumn{1}{c|}{HAIR \cite{cao2024hair}} & \multicolumn{1}{c|}{46.45 / 0.988} & \multicolumn{1}{c|}{37.93 / 0.919} & \multicolumn{1}{c|}{42.77 / 0.964} & \multicolumn{1}{c|}{HAIR \cite{cao2024hair}} & \multicolumn{1}{c|}{36.84 / 0.915} & \multicolumn{1}{c|}{31.47 / 0.745} & \multicolumn{1}{c}{32.54 / 0.796} \\
    \multicolumn{1}{c|}{} & \multicolumn{1}{c|}{InstructIR \cite{conde2024instructir}} & \multicolumn{1}{c|}{34.05 / 0.798} & \multicolumn{1}{c|}{27.03 / 0.602} & \multicolumn{1}{c|}{32.68 / 0.766} & \multicolumn{1}{c|}{InstructIR \cite{conde2024instructir}} & \multicolumn{1}{c|}{36.51 / 0.913} & \multicolumn{1}{c|}{31.34 / 0.741} & \multicolumn{1}{c}{32.52 / 0.793} \\
    \multicolumn{1}{c|}{} & \multicolumn{1}{c|}{PromptHSI \cite{lee2024prompthsi}} & \multicolumn{1}{c|}{32.52 / 0.817} & \multicolumn{1}{c|}{36.73 / 0.912} & \multicolumn{1}{c|}{39.75 / 0.954} & \multicolumn{1}{c|}{PromptHSI \cite{lee2024prompthsi}} & \multicolumn{1}{c|}{35.03 / 0.904} & \multicolumn{1}{c|}{30.75 / 0.732} & \multicolumn{1}{c}{31.85 / 0.775} \\
    \multicolumn{1}{c|}{} & \multicolumn{1}{c|}{  MP-HSIR (Ours)} & \multicolumn{1}{c|}{\textcolor{red}{48.07} / \textcolor{red}{0.990}} & \multicolumn{1}{c|}{\textcolor{blue}{39.00} / \textcolor{blue}{0.927}} & \multicolumn{1}{c|}{\textcolor{blue}{43.72} / \textcolor{blue}{0.968}} & \multicolumn{1}{c|}{  MP-HSIR (Ours)} & \multicolumn{1}{c|}{\textcolor{blue}{38.25} / \textcolor{blue}{0.924}} & \multicolumn{1}{c|}{\textcolor{red}{31.97} / \textcolor{red}{0.760}} & \multicolumn{1}{c}{\textcolor{red}{33.06} / \textcolor{red}{0.810}}   \\
    \hline
    \addlinespace[0.2em] 
    \hline
    \multicolumn{1}{c|}{\multirow{3}{*}{\textbf{Method}}} & \multicolumn{2}{c|}{\textbf{\textit{Inpainting (Mask Rate = 0.7, 0.8, 0.9)}}} & \multicolumn{1}{c|}{\multirow{3}{*}{\textbf{Method}}} & \multicolumn{2}{c|}{\textbf{\textit{Dehazing (Omega = 0.5, 0.75, 1)}}} & \multicolumn{1}{c|}{\multirow{3}{*}{\textbf{Method}}} & \multicolumn{2}{c}{\textbf{\textit{Completion (Rate = 0.1, 0.2 0.3)}}}  \\
\cline{2-3}\cline{5-6}\cline{8-9}    \multicolumn{1}{c|}{} & \multicolumn{1}{c|}{ICVL \cite{arad2016sparse}} & \multicolumn{1}{c|}{Chikusei \cite{yokoya2016airborne}} & \multicolumn{1}{c|}{} & \multicolumn{1}{c|}{PaviaU \cite{huang2009comparative}} & \multicolumn{1}{c|}{Eagle \cite{peerbhay2013commercial}} & \multicolumn{1}{c|}{} & \multicolumn{1}{c|}{ARAD \cite{arad2022ntire}} & \multicolumn{1}{c}{Berlin \cite{okujeni2016berlin}}  \\
\cline{2-3}\cline{5-6}\cline{8-9}    \multicolumn{1}{c|}{} & \multicolumn{1}{c|}{PSNR / SSIM $\uparrow$} & \multicolumn{1}{c|}{PSNR / SSIM $\uparrow$} & \multicolumn{1}{c|}{} & \multicolumn{1}{c|}{PSNR / SSIM $\uparrow$} & \multicolumn{1}{c|}{PSNR / SSIM $\uparrow$} & \multicolumn{1}{c|}{} & \multicolumn{1}{c|}{PSNR / SSIM $\uparrow$} & \multicolumn{1}{c}{PSNR / SSIM $\uparrow$}  \\
    \hline
    \multicolumn{1}{c|}{NAFNet \cite{chen2022simple}} & \multicolumn{1}{c|}{44.39 / 0.987} & \multicolumn{1}{c|}{\textcolor{blue}{39.82} / \textcolor{blue}{0.953}} & \multicolumn{1}{c|}{SGNet \cite{ma2022spectral}} & \multicolumn{1}{c|}{34.28 / 0.963} & \multicolumn{1}{c|}{37.22 / 0.976} & \multicolumn{1}{c|}{NAFNet \cite{chen2022simple}} & \multicolumn{1}{c|}{47.04 / 0.995} & \multicolumn{1}{c}{38.35 / 0.907}   \\
    \multicolumn{1}{c|}{Restormer \cite{zamir2022restormer}} & \multicolumn{1}{c|}{45.79 / \textcolor{blue}{0.990}} & \multicolumn{1}{c|}{36.33 / 0.899} & \multicolumn{1}{c|}{SCANet \cite{guo2023scanet}} & \multicolumn{1}{c|}{36.59 / 0.978} & \multicolumn{1}{c|}{39.64 / 0.985} & \multicolumn{1}{c|}{Restormer \cite{zamir2022restormer}} & \multicolumn{1}{c|}{48.34 / 0.995} & \multicolumn{1}{c}{35.07 / 0.606} \\
    \multicolumn{1}{c|}{DDS2M \cite{miao2023dds2m}} & \multicolumn{1}{c|}{42.18 / 0.969} & \multicolumn{1}{c|}{34.94 / 0.887} & \multicolumn{1}{c|}{MB-Taylor \cite{qiu2023mb}} & \multicolumn{1}{c|}{\textcolor{blue}{37.99} / 0.983} & \multicolumn{1}{c|}{\textcolor{blue}{41.03} / 0.991} & \multicolumn{1}{c|}{SwinIR \cite{liang2021swinir}} & \multicolumn{1}{c|}{49.75 / 0.995} & \multicolumn{1}{c}{35.45 / 0.886} \\
    \multicolumn{1}{c|}{HIR-Diff \cite{pang2024hir}} & \multicolumn{1}{c|}{38.91 / 0.949} & \multicolumn{1}{c|}{37.65 / 0.916} & \multicolumn{1}{c|}{DCMPNet \cite{zhang2024depth}} & \multicolumn{1}{c|}{37.20 / \textcolor{blue}{0.985}} & \multicolumn{1}{c|}{40.24 / 0.990} & \multicolumn{1}{c|}{MambaIR \cite{guo2024mambair}} & \multicolumn{1}{c|}{50.26 / 0.995} & \multicolumn{1}{c}{36.12 / 0.888}   \\
    \hline
    \multicolumn{1}{c|}{AirNet \cite{li2022all}} & \multicolumn{1}{c|}{42.60 / 0.981} & \multicolumn{1}{c|}{37.46 / 0.919} & \multicolumn{1}{c|}{AirNet \cite{li2022all}} & \multicolumn{1}{c|}{35.59 / 0.965} & \multicolumn{1}{c|}{38.83 / 0.981} & \multicolumn{1}{c|}{AirNet \cite{li2022all}} & \multicolumn{1}{c|}{45.27 / 0.992} & \multicolumn{1}{c}{35.91 / 0.624}   \\
    \multicolumn{1}{c|}{PromptIR \cite{potlapalli2023promptir}} & \multicolumn{1}{c|}{\textcolor{blue}{46.38} / \textcolor{blue}{0.990}} & \multicolumn{1}{c|}{38.07 / 0.930} & \multicolumn{1}{c|}{PromptIR \cite{potlapalli2023promptir}} & \multicolumn{1}{c|}{37.41 / 0.982} & \multicolumn{1}{c|}{40.73 / \textcolor{blue}{0.992}} & \multicolumn{1}{c|}{PromptIR \cite{potlapalli2023promptir}} & \multicolumn{1}{c|}{46.60 / 0.994} & \multicolumn{1}{c}{40.45 / 0.652} \\
    \multicolumn{1}{c|}{PIP \cite{li2023prompt}} & \multicolumn{1}{c|}{43.37 / 0.982} & \multicolumn{1}{c|}{38.43 / 0.930} & \multicolumn{1}{c|}{PIP \cite{li2023prompt}} & \multicolumn{1}{c|}{37.62 / 0.982} & \multicolumn{1}{c|}{40.74 / 0.990} & \multicolumn{1}{c|}{PIP \cite{li2023prompt}} & \multicolumn{1}{c|}{47.36 / 0.993} & \multicolumn{1}{c}{37.20 / 0.668} \\
    \multicolumn{1}{c|}{HAIR \cite{cao2024hair}} & \multicolumn{1}{c|}{44.02 / 0.982} & \multicolumn{1}{c|}{38.05 / 0.927} & \multicolumn{1}{c|}{HAIR \cite{cao2024hair}} & \multicolumn{1}{c|}{36.76 / 0.978} & \multicolumn{1}{c|}{40.86 / \textcolor{blue}{0.992}} & \multicolumn{1}{c|}{HAIR \cite{cao2024hair}} & \multicolumn{1}{c|}{45.08 / 0.992} & \multicolumn{1}{c}{38.01 / 0.650} \\
    \multicolumn{1}{c|}{InstructIR \cite{conde2024instructir}} & \multicolumn{1}{c|}{44.07 / 0.986} & \multicolumn{1}{c|}{36.11 / 0.907} & \multicolumn{1}{c|}{InstructIR \cite{conde2024instructir}} & \multicolumn{1}{c|}{34.72 / 0.971} & \multicolumn{1}{c|}{37.65 / 0.982} & \multicolumn{1}{c|}{InstructIR \cite{conde2024instructir}} & \multicolumn{1}{c|}{\textcolor{blue}{51.31} / \textcolor{blue}{0.997}} & \multicolumn{1}{c}{35.97 / 0.580} \\
    \multicolumn{1}{c|}{PromptHSI \cite{lee2024prompthsi}} & \multicolumn{1}{c|}{41.48 / 0.972} & \multicolumn{1}{c|}{37.33 / 0.946} & \multicolumn{1}{c|}{PromptHSI \cite{lee2024prompthsi}} & \multicolumn{1}{c|}{36.77 / 0.974} & \multicolumn{1}{c|}{39.78 / 0.984} & \multicolumn{1}{c|}{PromptHSI \cite{lee2024prompthsi}} & \multicolumn{1}{c|}{47.34 / 0.994} & \multicolumn{1}{c}{\textcolor{blue}{43.36} / \textcolor{blue}{0.972}} \\
    \multicolumn{1}{c|}{MP-HSIR (Ours)} & \multicolumn{1}{c|}{\textcolor{red}{51.53} / \textcolor{red}{0.996}} & \multicolumn{1}{c|}{\textcolor{red}{43.63} / \textcolor{red}{0.979}} & \multicolumn{1}{c|}{  MP-HSIR (Ours)} & \multicolumn{1}{c|}{\textcolor{red}{39.59} / \textcolor{red}{0.986}} & \multicolumn{1}{c|}{\textcolor{red}{42.41} / \textcolor{red}{0.995}} & \multicolumn{1}{c|}{  MP-HSIR (Ours)} & \multicolumn{1}{c|}{\textcolor{red}{56.48} / \textcolor{red}{0.999}} & \multicolumn{1}{c}{\textcolor{red}{49.50} / \textcolor{red}{0.987}}   \\
    \hline
    \addlinespace[0.2em] 
    \toprule
    \end{tabular}%
    }
  \caption{\textbf{[All-in-one]} Quantitative comparison of all-in-one and state-of-the-art task-specific methods on 7 HSI restoration tasks. The best and second-best performances are highlighted in \textcolor{red}{red} and \textcolor{blue}{blue}, respectively.}
  \vskip -3pt
  \label{tab:table1}%
\end{table*}%

\subsection{Experimental setup}

\noindent \textbf{Datasets.}
We employ 13 hyperspectral datasets for training and testing, comprising two natural scene HSI datasets and 11 remote sensing HSI datasets. The natural scene datasets include ICVL \cite{arad2016sparse} and ARAD \cite{arad2022ntire}, while the remote sensing datasets consist of Xiong’an \cite{yi2020aerial}, WDC \cite{zhu2017hyperspectral}, Pavia City and Pavia University \cite{huang2009comparative}, Houston \cite{wu2017convolutional}, Chikusei \cite{yokoya2016airborne}, Eagle \cite{peerbhay2013commercial}, Berlin \cite{okujeni2016berlin}, APEX \cite{itten2008apex}, Urban \cite{baumgardner2015indianpines}, and 10 EO-1 Hyperion scenes collected in this study. For the natural scene HSI datasets, 1,000 images are cropped into 64×64×31 patches for training, while an additional 100 non-overlapping images are reserved for testing. For the first eight remote sensing datasets, 80\% of each image is cropped into 64×64×100 patches for training, with the remaining regions used for testing. The last three remote sensing datasets are designated for real-world experiments. Due to the substantial domain differences between natural scenes and remote sensing data, we train models separately on each dataset category. To ensure a comprehensive evaluation, multiple test datasets are randomly assigned to each task, ensuring that both natural scene and remote sensing HSI datasets are included in every task. Moreover, normalization is conducted on all datasets. A detailed description of the datasets is provided in Section B of the supplementary material.

\noindent \textbf{Settings.}
To validate the effectiveness and generalization capability of the proposed model, we conducted evaluations under three different settings. \textbf{(1) All-in-one:} A unified model was trained on seven HSI restoration tasks, including Gaussian denoising, complex denoising, Gaussian deblurring, super-resolution, inpainting, dehazing, and band completion. \textbf{(2) Generalization:} Few-shot and zero-shot learning were performed on unseen tasks. Motion deblurring was evaluated after fine-tuning, while Poisson denoising was assessed directly using the pre-trained model. \textbf{(3) Real-World:} For real-world denoising, following the settings of \cite{li2023spatial}, the model was fine-tuned on the APEX dataset \cite{itten2008apex} and evaluated on the Urban dataset \cite{baumgardner2015indianpines}. For real-world dehazing, the model was directly evaluated on the EO-1 Hyperion data. Detailed experimental settings for each task, including degradation synthesis models and intensity ranges, are provided in Section C of the supplementary material.

\noindent \textbf{Implementation Details.}
Due to the difference in data volume, we employed distinct training strategies for natural scene and remote sensing HSI datasets. Both utilized the AdamW optimizer ($\beta_1=0.9, \beta_2=0.999$) with L1 loss and a batch size of 32. For natural HSIs, training was conducted for 100 epochs with an initial learning rate of $2 \times 10^{-4}$, which was gradually reduced to $1 \times 10^{-6}$ using cosine annealing \cite{loshchilov2016sgdr}. In contrast, for remote sensing HSIs, the initial learning rate was set to $1 \times 10^{-4}$, and training was extended to 300 epochs. Given the higher spectral dimensionality of remote sensing HSIs, the width of the all-in-one model was increased by a factor of 1.5, while task-specific models were fine-tuned to achieve optimal performance.

\begin{figure*}[tp]
    \centering
    \begin{minipage}{1.0\textwidth}
        \centering
        \includegraphics[width=1.0\textwidth]{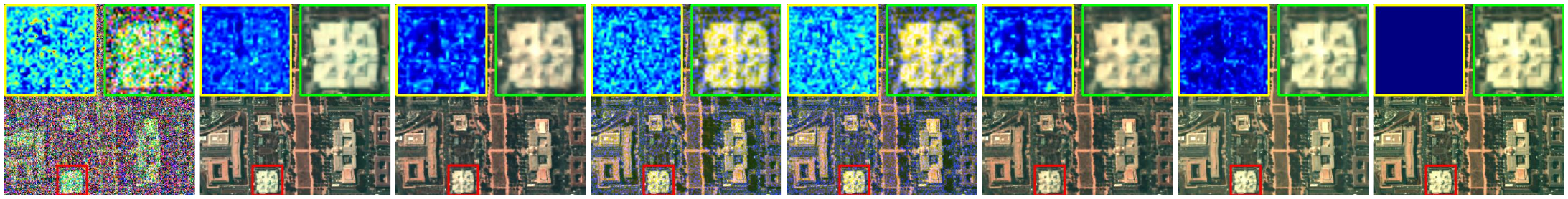}
        \vskip -7pt  
        \makebox[\textwidth][l]{\fontsize{6.5pt}{15pt}\selectfont \hspace{14pt} Input / 14.07 \hspace{19pt} SST \cite{li2023spatial} / 30.53 
        \hspace{11pt} LDERT \cite{li2024latent} / 30.74 \hspace{3pt} PromptIR \cite{potlapalli2023promptir} / 26.38
        \hspace{1pt} InstructIR \cite{conde2024instructir} / 24.63 \hspace{0pt} PromptHSI \cite{lee2024prompthsi} / 30.78  
        \hspace{12pt} Ours / 31.36 \hspace{21pt} GT / PSNR (dB)}
        \vskip 10pt  
    \end{minipage}
    
    \vspace{0pt}

    \begin{minipage}{1.0\textwidth}
        \centering
        \includegraphics[width=1.0\textwidth]{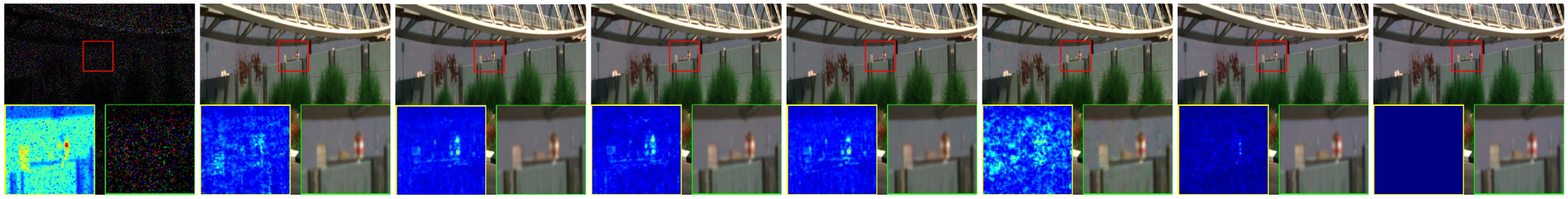}
        \vskip -7pt  
        \makebox[\textwidth][l]{\fontsize{6.5pt}{15pt}\selectfont \hspace{14pt} Input / 14.20 \hspace{15pt} NAFNet \cite{chen2022simple} / 40.82 
        \hspace{4pt} Restormer \cite{zamir2022restormer} / 42.16 \hspace{1pt} PromptIR \cite{potlapalli2023promptir} / 42.65 
        \hspace{1pt} InstructIR \cite{conde2024instructir} / 41.43 \hspace{0pt} PromptHSI \cite{lee2024prompthsi} / 38.52  
        \hspace{12pt} Ours / 47.67 \hspace{21pt} GT / PSNR (dB)}
        \vskip 10pt  
    \end{minipage}

    \vspace{0pt}
    
    \begin{minipage}{1.0\textwidth}
        \centering
        \includegraphics[width=1.0\textwidth]{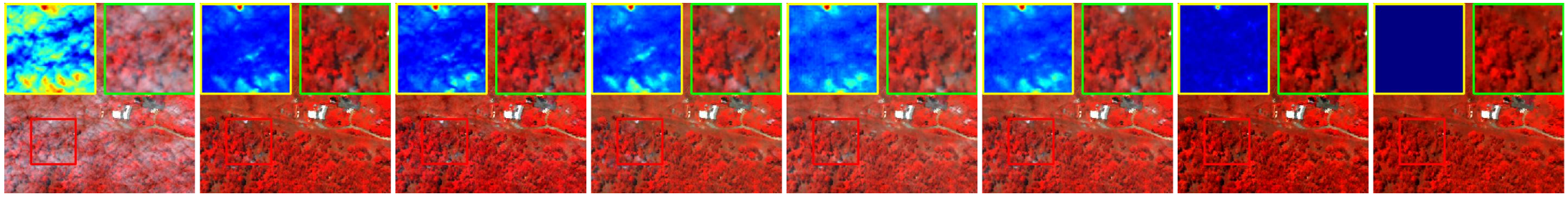}
        \vskip -7pt  
        \makebox[\textwidth][l]{\fontsize{6.5pt}{15pt}\selectfont \hspace{14pt} Input / 18.24 \hspace{9pt} MB-Taylor \cite{qiu2023mb} / 38.34 
        \hspace{0pt} DCMPNet \cite{zhang2024depth} / 37.68 \hspace{0pt} PromptIR \cite{potlapalli2023promptir} / 37.35 
        \hspace{0pt} InstructIR \cite{conde2024instructir} / 33.76 \hspace{0pt} PromptHSI \cite{lee2024prompthsi} / 37.82 
        \hspace{12pt} Ours / 39.49 \hspace{21pt} GT / PSNR (dB)}
    \end{minipage}
    \vskip -5pt
    \caption{\textbf{[All-in-one]} Visual comparison results on \textbf{\textit{complex denoising}}, \textbf{\textit{inpainting}}, and \textbf{\textit{dehazing}}. The zoomed-in regions are highlighted with \textcolor{red}{red} boxes, the corresponding zoomed results are marked with \textcolor{green}{green} boxes, and the residual maps are outlined with \textcolor{yellow}{yellow} boxes.}
    \label{fig:4}
\end{figure*}

\begin{figure*}[tp]
    \centering
    \includegraphics[width=1.0\textwidth]{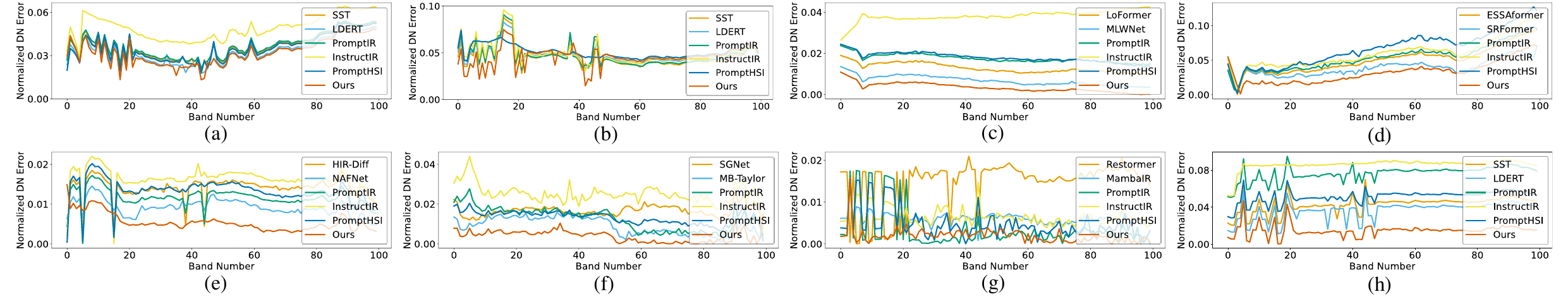} 
    \vskip -5pt
    \caption{The spectral errors of the restored HSIs across multiple tasks. (a) \textbf{\textit{Gaussian denoising}}. (b) \textbf{\textit{Complex denoising}}. (c) \textbf{\textit{Gaussian deblurring}}. (d) \textbf{\textit{Super-resolution}}. (e) \textbf{\textit{Inpainting}}. (f) \textbf{\textit{Dehazing}}. (g) \textbf{\textit{Band completion}}. (h) \textbf{\textit{Poisson denoising}}. Zoom in for the best view.}
    \vskip -5pt
    \label{fig:5}
\end{figure*}

\begin{table}[t]
  \centering
  \renewcommand{\arraystretch}{1.2} 
  \resizebox{\columnwidth}{!}{
    \begin{tabular}{c|c|cc|c|cc}
    \hline
    \multirow{2}{*}{\textbf{Type}} & \multirow{2}{*}{\textbf{Method}} & \multicolumn{2}{c|}{\textbf{\textit{Motion Deblurring}}} & \multirow{2}{*}{\textbf{Methods}} & \multicolumn{2}{c}{\textbf{\textit{Poisson Denoising}}} \\
\cline{3-4}\cline{6-7}          &       & \multicolumn{1}{c}{\hspace{3pt}PSNR $\uparrow$}  & \multicolumn{1}{c|}{\hspace{3pt}SSIM $\uparrow$}  &       & \multicolumn{1}{c}{\hspace{3pt}PSNR $\uparrow$}  & \multicolumn{1}{c}{\hspace{3pt}SSIM $\uparrow$} \\
    \hline
    \multirow{4}[2]{*}{\makecell{Task\\Specific}} & Stripformer \cite{tsai2022stripformer} & \hspace{3pt}37.91  & \hspace{3pt}0.948 & QRNN3D \cite{wei20203} & \hspace{3pt}35.98 & \hspace{3pt}0.910 \\
          & FFTformer \cite{kong2023efficient} & \hspace{3pt}38.42 & \hspace{3pt}0.950 & SST \cite{li2023spatial}   & \hspace{3pt}37.51 & \hspace{3pt}0.914 \\
          & LoFormer \cite{mao2024loformer} & \hspace{3pt}38.94 & \hspace{3pt}0.953 & SERT \cite{li2023spectral}  & \hspace{3pt}37.46 &  \hspace{3pt}0.913 \\
          & MLWNet \cite{gao2024efficient} & \hspace{3pt}\textcolor{blue}{39.28} & \hspace{3pt}\textcolor{blue}{0.959} & LDERT \cite{li2024latent} & \hspace{3pt}\textcolor{blue}{37.92} & \hspace{3pt}\textcolor{blue}{0.915} \\
    \hline
    \multirow{7}[2]{*}{\makecell{All\\in\\One}} & AirNet \cite{li2022all} & \hspace{3pt}37.43 & \hspace{3pt}0.947 & AirNet \cite{li2022all} & \hspace{3pt}33.91 & \hspace{3pt}0.896 \\
          & PromptIR \cite{potlapalli2023promptir} & \hspace{3pt}38.77 & \hspace{3pt}0.951 & PromptIR \cite{potlapalli2023promptir} & \hspace{3pt}34.83  & \hspace{3pt}0.902 \\
          & PIP \cite{li2023prompt}   & \hspace{3pt}38.55 & \hspace{3pt}0.950 & PIP \cite{li2023prompt}   & \hspace{3pt}34.52 & \hspace{3pt}0.901 \\
          & HAIR \cite{cao2024hair}  & \hspace{3pt}37.58 & \hspace{3pt}0.948 & HAIR \cite{cao2024hair}  & \hspace{3pt}34.38 & \hspace{3pt}0.901 \\
          & InstructIR \cite{conde2024instructir} & \hspace{3pt}36.58 & \hspace{3pt}0.942 & InstructIR \cite{conde2024instructir} & \hspace{3pt}33.96 & \hspace{3pt}0.895 \\
          & PromptHSI \cite{lee2024prompthsi} & \hspace{3pt}36.79 & \hspace{3pt}0.945 & PromptHSI \cite{lee2024prompthsi} & \hspace{3pt}37.25 & \hspace{3pt}0.912 \\
          & MP-HSIR (Ours)  & \hspace{3pt}\textcolor{red}{40.06} & \hspace{3pt}\textcolor{red}{0.965} & MP-HSIR (Ours)  & \hspace{3pt}\textcolor{red}{38.56} & \hspace{3pt}\textcolor{red}{0.922} \\
    \hline
    \end{tabular}%
    }
  \caption{\textbf{[Generalization]} The results of \textbf{\textit{motion deblurring}} on the ICVL dataset \cite{arad2016sparse} and \textbf{\textit{Poisson denoising}} on the Houston dataset \cite{wu2017convolutional}. Task-specific methods are trained on the entire dataset, while all-in-one methods are fine-tuned using only 5\% of the data.}
  \vskip -6pt
  \label{tab:table2}%
\end{table}%

\noindent \textbf{Evaluation Metrics and Comparisons Methods.}
We employed PSNR and SSIM \cite{wang2004image} as quantitative metrics for evaluation. The comparison methods included six all-in-one models, \textit{i.e.}, AirNet \cite{li2022all}, PromptIR \cite{potlapalli2023promptir}, PIP \cite{li2023prompt}, HAIR \cite{cao2024hair}, InstructIR \cite{conde2024instructir}, and PromptHSI \cite{lee2024prompthsi}, along with 22 state-of-the-art task-specific models.

\subsection{Comparison with state-of-the-art methods}

\noindent \textbf{All-in-one.}
Table \ref{tab:table1} presents the quantitative performance of seven all-in-one methods and 22 task-specific approaches across seven HSI restoration tasks. Our method consistently outperforms all all-in-one methods and surpasses state-of-the-art task-specific approaches in multiple tasks, achieving significant PSNR improvements in inpainting, dehazing, and band completion. Furthermore, our method attains the best or near-best performance across all hyperspectral remote sensing datasets. Notably, InstructIR and PromptHSI exhibit anomalous results in the deblurring task due to their inability to effectively distinguish different types of HSI degradation. Visualization comparison of prompts is provided in Section D of the supplementary material.

Figure \ref{fig:4} presents visual comparisons for complex denoising, inpainting, and dehazing. The residual maps highlight the superiority of our method in structural recovery and fine-detail reconstruction. Figure \ref{fig:5} illustrates the normalized digital number (DN) error between the restored and ground truth images across spectral bands. Leveraging spectral prompts, our method achieves the most accurate spectral reconstruction across all tasks.

\begin{figure*}[t]
    \centering
    \begin{minipage}{1.0\textwidth}
        \centering
        \includegraphics[width=1.0\textwidth]{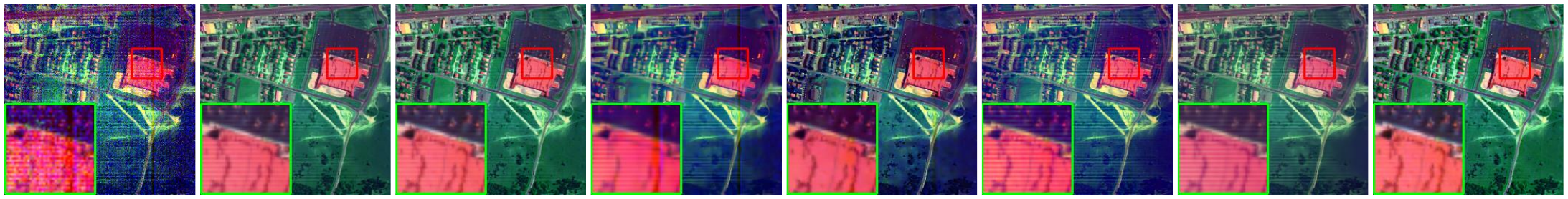}
        \vskip -6pt  
        \makebox[\textwidth][l]{\fontsize{6.5pt}{15pt}\selectfont \hspace{22pt} Input \hspace{42pt} SST \cite{li2023spatial}
        \hspace{31pt} LDERT \cite{li2024latent}  \hspace{28pt} AirNet \cite{li2022all} \hspace{25pt} 
        PromptIR \cite{potlapalli2023promptir} 
        \hspace{19pt} InstructIR \cite{conde2024instructir}  \hspace{19pt} PromptHSI \cite{lee2024prompthsi}
        \hspace{30pt} Ours}
        \vskip 10pt  
    \end{minipage}
    
    \vspace{0pt}

    \begin{minipage}{1.0\textwidth}
        \centering
        \includegraphics[width=1.0\textwidth]{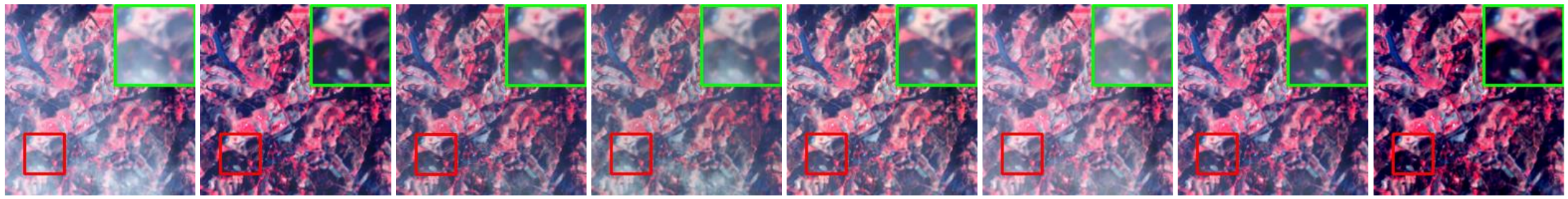}
        \vskip -6pt  
        \makebox[\textwidth][l]{\fontsize{6.5pt}{15pt}\selectfont \hspace{22pt} Input \hspace{32pt} MB-Taylor \cite{qiu2023mb}
        \hspace{19pt} DCMPNet \cite{zhang2024depth}  \hspace{24pt} AirNet \cite{li2022all}
        \hspace{25pt} PromptIR \cite{potlapalli2023promptir} 
        \hspace{19pt} InstructIR \cite{conde2024instructir}  \hspace{20pt} PromptHSI \cite{lee2024prompthsi}
        \hspace{30pt} Ours}
    \end{minipage}
    \vskip -5pt
    \caption{\textbf{[Real-world]} Visual comparison results on \textbf{\textit{real denoising}} and \textbf{\textit{real dehazing}}. Zoom in for the best view.}
    \vskip -10pt
    \label{fig:6}
\end{figure*}

\begin{table}[t]
  \centering
  \renewcommand{\arraystretch}{1.3} 
  \resizebox{\columnwidth}{!}{
    \begin{tabular}{lllccc}
    \hline
    \multicolumn{3}{l}{\textbf{Method}} & PSNR $\uparrow$  & SSIM $\uparrow$  & Params (M) \\
    \hline
    \multicolumn{3}{l}{Baseline (Only Spatial SA)} & 39.24  & 0.963  & 20.93  \\
    \hline
    \multicolumn{3}{l}{+ Textual Prompt $P_T$} & 39.62  & 0.964  & 21.51  \\
    \multicolumn{3}{l}{+ Visual Prompt $P_V$} & 39.57  & 0.964  & 23.68  \\
    \multicolumn{3}{l}{+ Textual Prompt $P_T$ + Visual Prompt $P_V$} & 39.90  & 0.964  & 24.26  \\
    \hline
    \multicolumn{3}{l}{+ Global Spectral SA + $P_T$ + $P_T$} & 40.63  & 0.969  & 30.07  \\
    \multicolumn{3}{l}{+ Local Spectral SA + $P_T$ + $P_V$} & 40.46  & 0.968  & 24.43  \\
    \multicolumn{3}{l}{+ Local Spectral SA + $P_T$ + $P_V$ + Spectral Prompt $P_S$} & 41.05  & 0.971  & 25.10  \\
    \hline
    \multicolumn{3}{l}{Full Model} & \textbf{41.98}  & \textbf{0.974}  & 30.91  \\
    \hline
    \end{tabular}%
    }
  \caption{Ablation study to verify the effectiveness of modules on Chikusei dataset in \textbf{\textit{inpainting}} task with mask ratio = 0.9.}
  \label{tab:table3_2}%
\end{table}%

\noindent \textbf{Generalization.}
Table \ref{tab:table2} presents the experimental results for motion blur removal using few-shot learning and Poisson noise removal using zero-shot learning. The proposed method exhibits superior generalization ability compared to existing approaches under both learning paradigms.

\noindent \textbf{Real-world.}
As shown in Figure \ref{fig:6}, we present a comparative visualization of all-in-one and task-specific methods applied to real-world denoising and dehazing scenarios. The results highlight the proposed method's effectiveness in restoring texture details and spectral structure, while mitigating common artifacts like over-smoothing and spectral distortion. 

\noindent \textbf{Additional Results.}
More quantitative results and visualizations are in Section E of the supplementary material.

\subsection{Model Analysis}
\noindent \textbf{Ablation Study.} To assess the effectiveness of each module in the proposed MP-HSIR framework, we conducted an ablation study, starting with a baseline model that includes only spatial self-attention (SA) and then progressively incorporating textual prompts $P_T$, learnable visual prompts $P_V$, global spectral self-attention, local spectral self-attention, and spectral prompts $P_S$. As shown in Table \ref{tab:table3_2}, both accuracy metrics for the inpainting task on the Chikusei dataset progressively improve with the addition of each module. Notably, the local spectral self-attention module guided by spectral prompts provides the most significant performance gain, attributed to the additional spectral guidance offered by spectral prompts. Results for other tasks are provided in Section E of the supplementary material.

\noindent \textbf{Visualization of Spectral Prompt.} To further investigate the role of spectral prompts, we visualize the activation distribution of universal low-rank spectral patterns under both noise and blur degradations in Figure \ref{fig:7}. By selecting identical regions from both degraded images for comparison, we observe that despite different degradation types, the activation patterns remain highly consistent within the same regions. This indicates that spectral prompts effectively capture key spectral information in HSI and demonstrate strong robustness against different types of degradation.

\begin{figure}[t]
    \centering
    \includegraphics[width=0.48\textwidth]{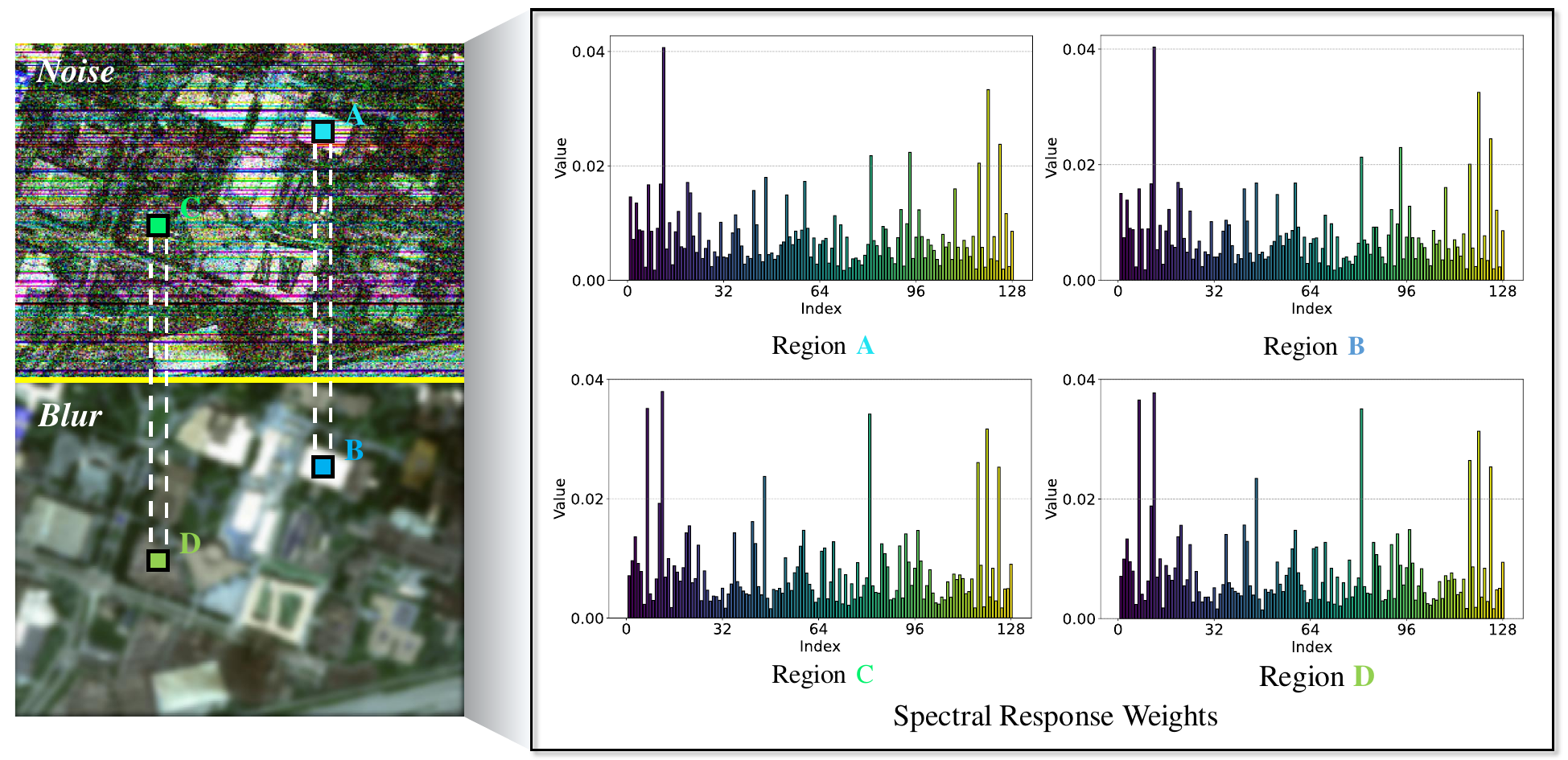} 
    \vskip -10pt
    \caption{Visualization of the activation proportions for universal low-rank spectral patterns within the spectral prompt.}
    \label{fig:7}
\end{figure}

\section{Conclusion}
\label{sec:conclusion}

HSI restoration is challenging due to diverse and often unknown degradation types, while existing methods are typically limited to specific degradation scenarios. In this work, we propose MP-HSIR, a universal HSI restoration framework that integrates spectral, textual, and visual prompts. MP-HSIR introduces a novel prompt-guided spatial-spectral transformer, where spectral prompts provide universal low-rank spectral patterns to enhance spectral reconstruction. Furthermore, the text-visual synergistic prompts combine high-level semantic representations with fine-grained visual features to guide degradation modeling. Extensive experiments on 9 HSI restoration tasks—including all-in-one restoration, generalization testing, and real-world scenarios—demonstrate the superior robustness and adaptability of MP-HSIR under diverse degradation conditions.

{
    \small
    \bibliographystyle{ieeenat_fullname}
    \bibliography{main}
}

\clearpage  
\twocolumn[  
    \begin{center}
        \fontsize{18pt}{24pt}\selectfont 
        \textbf{Supplementary Materials}
    \end{center}
    \vspace{1cm}
]

\renewcommand{\thesection}{\Alph{section}} 

\section{Degradation Preictor}
The proposed universal multi-prompt framework, MP-HSIR, can perform specific restoration tasks either based on human instructions or autonomously via the degradation predictor. In this work, we employ a ResNet-34 network with fast Fourier convolution \cite{chi2020fast} to train classification models separately on natural scene and remote sensing hyperspectral datasets. The binary cross-entropy loss used for training is defined as follows:

\begin{equation}
\label{eq1}
\mathcal{L}=-\frac{1}{N} \sum_{i=1}^N\left[y_i \log p_i+\left(1-y_i\right) \log \left(1-p_i\right)\right],
\end{equation}
where \( N \) denotes the number of samples, \( p_i \) represents the output after sigmoid activation, and \( y_i \) is the corresponding ground-truth label. The training was conducted with a batch size of 64, using the same optimizer as in the restoration experiments. The initial learning rate was set to \( 1 \times 10^{-4} \) and progressively decreased to \( 1 \times 10^{-6} \) via cosine annealing \cite{loshchilov2016sgdr}. The model was trained for 1000 epochs on the natural scene hyperspectral dataset and 4000 epochs on the remote sensing hyperspectral dataset.

Table \ref{tab:table13} presents the accuracy and precision of the degradation predictor for both hyperspectral datasets. The results show that the predictor achieves 100\% accuracy and precision across all tasks, demonstrating MPIR-HSI's effectiveness in supporting all trained blind restoration tasks.

\section{Dataset Details}
This section provides a comprehensive overview of the 13 datasets used across 9 hyperspectral image (HSI) restoration tasks and real-world scenarios, as summarized in Table~\ref{tab:table4}.

\noindent \textbf{ARAD \cite{arad2022ntire}.} The ARAD dataset, derived from the NTIRE 2022 Spectral Recovery Challenge, was collected using a Specim IQ hyperspectral camera. It consists of 1,000 images, with 900 allocated for training and 50 for testing.

\noindent \textbf{ICVL \cite{arad2016sparse}.} The ICVL dataset was obtained using a Specim PS Kappa DX4 hyperspectral camera combined with a rotating stage for spatial scanning. It contains 201 images, with 100 for training and 50 for testing, ensuring no scene overlap.

\noindent \textbf{Xiong’an \cite{yi2020aerial}.} The Xiong’an dataset was captured using an imaging spectrometer developed by the Chinese Academy of Sciences. Three central regions of size 512 × 512 were randomly cropped for testing, while the remaining areas were used for training.

\begin{table}[!t]
  \centering
  \renewcommand{\arraystretch}{1.1} 
  \resizebox{\columnwidth}{!}{
    \begin{tabular}{ccccc}
    \hline
    \multirow{2}{*}{\textbf{Task}} & \multicolumn{2}{c}{\textbf{Natural Scene}} & \multicolumn{2}{c}{\textbf{Remote Sensing}} \\
    \cline{2-5} 
    & Accuracy $\uparrow$ & Precision $\uparrow$ & Accuracy $\uparrow$ & Precision $\uparrow$ \\
    \hline
    Gaussian Denoising   & 100.00  & 100.00  & 100.00  & 100.00  \\
    Complex Denoising    & 100.00  & 100.00  & 100.00  & 100.00  \\
    Gaussian Deblurring  & 100.00  & 100.00  & 100.00  & 100.00  \\
    Super-Resolution     & 100.00  & 100.00  & 100.00  & 100.00  \\
    Inpainting          & 100.00  & 100.00  & 100.00  & 100.00  \\
    Dehazing            & 100.00  & 100.00  & 100.00  & 100.00  \\
    Band Completion     & 100.00  & 100.00  & 100.00  & 100.00  \\
    \hline
    \end{tabular}
  }
  \caption{Accuracy and precision results of the degradation predictor in degradation classification}
  \label{tab:table13}
\end{table}

\begin{table*}[htbp]
  \centering
  \renewcommand{\arraystretch}{0.6} 
  \tiny
  \resizebox{\textwidth}{!}{ 
    \begin{tabular}{ccccccc}
    \hline
       \textbf{Type} & \textbf{Dataset} & \textbf{Sensor} & \textbf{Wavelength (nm)} & \textbf{Channels} & \textbf{Size}  & \textbf{GSD (m)} \bigstrut\\
    \hline
    \multirow{2}[4]{*}{\textbf{\makecell{Natural\\HSIs\\\hspace{3mm}}
}} & ARAD\_1K & Specim IQ & 400–700 & 31    & 482×512 & / \bigstrut\\
\cline{2-7}          & ICVL  & Specim PS Kappa DX4 & 400–700 & 31    & 1392×1300 & / \bigstrut\\
    \hline
    \multirow{10}[20]{*}{\textbf{\makecell{Remote\\Sensing\\HSIs}}} & Xiong’an & Unknown & 400–1000 & 256   & 3750×1580 & 0.5 \bigstrut\\
\cline{2-7}          & WDC   & Hydice & 400–2400 & 191   & 1208×307 & 5 \bigstrut\\
\cline{2-7}          & PaviaC & ROSIS & 430–860 & 102   & 1096×715 & 1.3 \bigstrut\\
\cline{2-7}          & PaviaU & ROSIS & 430–860 & 103   & 610×340 & 1.3 \bigstrut\\
\cline{2-7}          & Houston & ITRES CASI-1500 & 364–1046 & 144   & 349×1905 & \textcolor[rgb]{ .251,  .251,  .251}{2.5} \bigstrut\\
\cline{2-7}          & Chikusei & HH-VNIR-C & 343–1018 & 128   & 2517×2335 & \textcolor[rgb]{ .251,  .251,  .251}{2.5} \bigstrut\\
\cline{2-7}          & Eagle & AsiaEAGLE II & 401–999 & 128   & 2082×1606 & \textcolor[rgb]{ .251,  .251,  .251}{1} \bigstrut\\
\cline{2-7}          & Berlin & Unknown & 455–2447 & 111   & 6805×1830 & 3.6 \bigstrut\\
\cline{2-7}          & Urban & Hydice & 400-2500 & 210   & 307×307 & 2 \bigstrut\\
\cline{2-7}          & APEX & Unknown & 350-2500 & 285   & 1000×1500 & 2 \bigstrut\\
\cline{2-7}          & EO-1  & Hyperion & 357-2567 & 242   & 3471×991 & 30 \bigstrut\\
    \hline
    \end{tabular}%
    }
  \caption{Properties of 13 Natural Scene and Remote Sensing Hyperspectral Datasets.}
  \label{tab:table4}%
\end{table*}%

\noindent \textbf{WDC \cite{zhu2017hyperspectral}.} The Washington DC (WDC) dataset was captured by a Hydice sensor. A central region of size 256 × 256 was selected for testing, with the remainder used for training.

\noindent \textbf{PaviaC \cite{huang2009comparative}.} The Pavia Center (PaviaC) dataset was acquired using a ROSIS sensor, following the same partitioning strategy as the WDC dataset.

\noindent \textbf{PaviaU \cite{huang2009comparative}.} The Pavia University (PaviaU) dataset was also collected using a ROSIS sensor, with the same partitioning strategy as WDC.

\noindent \textbf{Houston \cite{wu2017convolutional}.} The Houston dataset was obtained using an ITRES CASI-1500 sensor, employing the same partitioning strategy as WDC.

\noindent \textbf{Chikusei \cite{yokoya2016airborne}.} The Chikusei dataset was captured using an HH-VNIR-C sensor. Four 512 × 512 regions were randomly cropped for testing, with the remaining areas used for training.

\noindent \textbf{Eagle \cite{peerbhay2013commercial}.} The Eagle dataset was collected using an AsiaEAGLE II sensor, following the same partitioning strategy as WDC.

\noindent \textbf{Berlin \cite{okujeni2016berlin}.} The Berlin dataset utilizes only the HyMap image from the BerlinUrbGrad dataset. A 512 × 512 central region was randomly cropped for testing, while the remaining data were used for training.

\noindent \textbf{Urban \cite{baumgardner2015indianpines}.} The Urban dataset, collected using a Hydice sensor, is specifically used for real-world denoising experiments.

\noindent \textbf{APEX \cite{itten2008apex}.} The APEX dataset exhibits characteristics similar to the Urban dataset and is primarily used for fine-tuning pre-trained models.

\noindent \textbf{EO-1 \cite{folkman2001eo}.} The EO-1 dataset was captured by the Hyperion sensor. Ten scenes were collected for testing, with 67 invalid bands removed, retaining 175 valid bands for real-world dehazing experiments.

All datasets underwent min-max normalization, and training samples were uniformly cropped to 64 × 64.

\section{Detailed Experimental Setup}
In this section, we provide a detailed description of the experimental settings for the 9 HSI restoration tasks.

\textbf{Gaussian Denoising.} Each image was corrupted by zero-mean independent and identically distributed (i.i.d.) Gaussian noise with sigma ranging from 30 to 70. For testing, sigma = 30, 50, and 70 were selected for evaluation.

\textbf{Complex Denoising.} Each image was corrupted with one of the following four noise scenarios:

1) Case 1 (\textit{Non-i.i.d. Gaussian Noise}): All bands were corrupted by non-i.i.d. Gaussian noise with standard deviations randomly selected from 10 to 70.

2) Case 2 (\textit{Gaussian Noise} + \textit{Stripe Noise}): All bands were corrupted by non-i.i.d. Gaussian noise, and one-third of the bands were randomly selected to add column stripe noise with intensities ranging from 5\% to 15\%.

3) Case 3 (\textit{Gaussian Noise} + \textit{Deadline Noise}): The noise generation process was similar to Case 2, but stripe noise was replaced by deadline noise.

4) Case 4 (\textit{Gaussian Noise} + \textit{Impulse Noise}): All bands were corrupted by non-i.i.d. Gaussian noise, and one-third of the bands were randomly selected to add impulse noise with intensities ranging from 10\% to 70\%.

\textbf{Gaussian Deblurring.} An empirical formula was used to calculate the standard deviation $\sigma$ based on the Gaussian kernel size $K_{S}$, formulated as:

\begin{equation}
\label{eq2}
\sigma=0.3 \times\left(\frac{K_{S} -1}{2}-1\right)+0.8.
\end{equation}

For natural hyperspectral datasets, $K_{S}$ was set to 9, 15, and 21, while for remote sensing hyperspectral datasets, $K_{S}$ was set to 7, 11, and 15.

\textbf{Super-Resolution.} Bicubic interpolation was used to downsample the images, with downscaling factors of 2, 4, and 8. To ensure that the input and output image sizes of the all-in-one model remained consistent, an unpooling operation was applied to resize the downsampled HSIs to their original dimensions.

\textbf{Inpainting.} Random masks with rates of 0.7, 0.8, and 0.9 were applied to each image for the inpainting task.

\textbf{Dehazing.} To realistically simulate haze contamination, the haze synthesis method from \cite{guo2020rsdehazenet} was adopted. Specifically, 100 haze masks were extracted from the cirrus band of Landsat-8 OLI and superimposed onto the original image according to the wavelength ratio to generate haze-affected HSIs, modeled as:

\begin{equation}
\label{eq3}
I_i=J_i e^{\left(\frac{\lambda_1}{\lambda_i}\right)^\gamma \ln {t}_1}+A\left(1-e^{\left(\frac{\lambda_1}{\lambda_i}\right)^\gamma \ln {t}_1}\right),
\end{equation}
where $I$ is the hazy HSI, $J$ is the clear HSI, $A$ is the global atmospheric light, $\lambda$ is the wavelength, and $\gamma$ is the spatial function, which
is set to 1. The reference transmission map ${t}_1$ is calculated from the cirrus band reflectance:

\begin{equation}
\label{eq4}
t_1=1-\omega B_9,
\end{equation}
where $\omega$ is a weighting factor controlling the haze intensity, and $B_9$ is the cirrus band reflectance. In the experiments, $\omega$ was set to [0.5, 0.75, 1], corresponding to different levels of haze contamination.

\textbf{Band Completion.} A certain proportion of bands were discarded for each image, with discard rates of 0.1, 0.2, and 0.3. The experimental results were evaluated only on the missing bands.

\textbf{Motion Deblurring.} The pre-trained model was fine-tuned and tested on this task with a blur kernel radius of 15 and a blur angle of 45 degrees.

\textbf{Poisson Denoising.} The pre-trained model was directly tested on this task with a Poisson noise intensity scaling factor of 10.



\begin{table*}[!t]
  \centering
  \renewcommand{\arraystretch}{1.2} 
  \resizebox{\textwidth}{!}{
    \begin{tabular}{c|c|c|c|c|c|c|c|c|c|c}
\cline{1-11}    \multirow{3}[6]{*}{\textbf{Type}} & \multirow{3}[6]{*}{\textbf{Methods}} & \multicolumn{3}{c|}{\textbf{\textit{Gaussian Denoising} (ICVL \cite{arad2016sparse})}} & \multicolumn{3}{c|}{\textbf{\textit{Gaussian Denoising} (ARAD \cite{arad2022ntire})}} & \multicolumn{3}{c}{\textbf{\textit{Gaussian Denoising} (Xiong’an \cite{yi2020aerial})}} \bigstrut\\
\cline{3-11}          &       & Sigma = 30 & Sigma = 50 & Sigma = 70 & Sigma = 30 & Sigma = 50 & Sigma = 70 & Sigma = 30 & Sigma = 50 & Sigma = 70 \bigstrut\\
\cline{3-11}          &       & PSNR / SSIM $\uparrow$& PSNR / SSIM $\uparrow$& PSNR / SSIM $\uparrow$& PSNR / SSIM $\uparrow$& PSNR / SSIM $\uparrow$& PSNR / SSIM $\uparrow$& PSNR / SSIM $\uparrow$& PSNR / SSIM $\uparrow$& PSNR / SSIM $\uparrow$\bigstrut\\
    \hline
    \multirow{4}[2]{*}{\makecell{Task\\Specific}} & QRNN3D \cite{wei20203}& 42.18 / 0.967 & 39.70 / 0.942 & 38.09 / 0.933 & 41.67 / 0.967 & 39.15 / 0.935 & 36.71 / 0.894 & 37.86 / 0.870 & 36.03 / 0.825 & 34.29 / 0.792 \bigstrut[t]\\
          & SST \cite{li2023spatial}   & 43.32 / 0.976 & 41.09 / 0.952 & 39.51 / 0.949 & 43.02 / 0.972 & 40.58 / 0.951 & 38.99 / 0.941 & 39.26 / 0.878 & 37.34 / 0.848 & 35.99 / 0.824 \\
          & SERT \cite{li2023spectral} & 43.53 / \textcolor{blue}{0.978} & 41.32 / \textcolor{blue}{0.966} & \textcolor{blue}{39.82} / \textcolor{blue}{0.956} & \textcolor{blue}{43.21} / \textcolor{blue}{0.975} & 40.84 / 0.959 & 39.21 / 0.945 & 39.54 / 0.885 & 37.58 / 0.859 & 36.37 / 0.833 \\
          & LDERT \cite{li2024latent}& \textcolor{red}{44.12} / \textcolor{red}{0.982} & \textcolor{red}{41.68} / \textcolor{red}{0.968} & \textcolor{red}{39.95} / \textcolor{red}{0.957} & \textcolor{red}{43.74} / \textcolor{red}{0.979} & \textcolor{red}{41.35} / \textcolor{red}{0.966} & \textcolor{red}{39.32} / \textcolor{red}{0.950} & \textcolor{blue}{39.92} / 0.889 & \textcolor{blue}{37.96} / 0.868 & 36.54 / 0.838 \bigstrut[b]\\
    \hline
    \multirow{7}[2]{*}{\makecell{All\\in\\One}} & AirNet \cite{li2022all}& 42.02 / 0.966 & 39.68 / 0.942 & 37.59 / 0.923 & 41.39 / 0.963 & 39.08 / 0.933 & 37.09 / 0.903 & 34.04 / 0.700 & 31.61 / 0.665 & 30.17 / 0.639 \bigstrut[t]\\
          & PromptIR \cite{potlapalli2023promptir}& 42.40 / 0.971 & 40.14 / 0.954 & 38.20 / 0.934 & 41.84 / 0.967 & 39.55 / 0.947 & 37.67 / 0.921 & 34.90 / 0.715 & 32.76 / 0.680 & 31.31 / 0.657 \\
          & PIP \cite{li2023prompt}  & 43.00 / 0.974 & 40.69 / 0.958 & 38.94 / 0.941 & 42.33 / 0.970 & 40.07 / 0.953 & 38.36 / 0.933 & 34.51 / 0.704 & 32.43 / 0.671 & 30.98 / 0.647 \\
          & HAIR \cite{cao2024hair} & 42.53 / 0.972 & 40.23 / 0.957 & 38.78 / 0.939 & 42.03 / 0.968 & 39.76 / 0.950 & 37.95 / 0.928 & 34.51 / 0.712 & 32.22 / 0.675 & 30.89 / 0.650 \\
          & InstructIR \cite{conde2024instructir} & 42.99 / 0.974 & 40.84 / 0.960 & 39.23 / 0.946 & 42.21 / 0.970 & 40.16 / 0.955 & 38.60 / 0.938 & 33.79 / 0.703 & 31.47 / 0.662 & 29.96 / 0.633 \\
          & PromptHSI \cite{lee2024prompthsi} & 42.61 / 0.976 & 40.27 / 0.960 & 39.08 / 0.945 & 41.90 / 0.971 & 39.84 / 0.959 & 38.37 / 0.938 & 39.54 / \textcolor{blue}{0.902} & 37.80 / \textcolor{blue}{0.877} & \textcolor{blue}{36.87} / \textcolor{blue}{0.864} \\
          & MP-HSIR (Ours)  & \textcolor{blue}{43.62} / 0.977 & \textcolor{blue}{41.41} / 0.963 & \textcolor{blue}{39.82} / 0.951 & 43.12 / \textcolor{blue}{0.975} & \textcolor{blue}{40.88} / \textcolor{blue}{0.960} & \textcolor{blue}{39.28} / \textcolor{blue}{0.946} & \textcolor{red}{40.55} / \textcolor{red}{0.922} & \textcolor{red}{38.70} / \textcolor{red}{0.896} & \textcolor{red}{37.17} / \textcolor{red}{0.874} \bigstrut[b]\\
    \hline
    \end{tabular}%
    }
  \caption{\textbf{[All-in-one]} Quantitative comparison of all-in-one and state-of-the-art task-specific methods under different Gaussian noise levels on \textbf{\textit{Gaussian denoising}} tasks. The best and second-best performances are highlighted in \textcolor{red}{red} and \textcolor{blue}{blue}, respectively.}
  \label{tab:table5}%
\end{table*}%

\begin{table*}[!t]
  \centering
  \renewcommand{\arraystretch}{1.2} 
  \resizebox{\textwidth}{!}{
    \begin{tabular}{c|c|c|c|c|c|c|c|c|c|c|c|c|c}
    \hline
    \multirow{3}[6]{*}{\textbf{Type}} & \multirow{3}[6]{*}{\textbf{Methods}} & \multicolumn{4}{c|}{\textbf{\textit{Complex Denoising} (ICVL \cite{arad2016sparse})}} & \multicolumn{4}{c|}{\textbf{\textit{Complex Denoising} (ARAD \cite{arad2022ntire})}} & \multicolumn{4}{c}{\textbf{\textit{Complex Denoising} (WDC \cite{zhu2017hyperspectral})}} \bigstrut\\
\cline{3-14}          &       & Case = 1 & Case = 2 & Case = 3 & Case = 4 & Case = 1 & Case = 2 & Case = 3 & Case = 4 & Case = 1 & Case = 2 & Case = 3 & Case = 4 \bigstrut\\
\cline{3-14}          &       & PSNR / SSIM $\uparrow$& PSNR / SSIM $\uparrow$& PSNR / SSIM $\uparrow$& PSNR / SSIM $\uparrow$& PSNR / SSIM $\uparrow$& PSNR / SSIM $\uparrow$& PSNR / SSIM $\uparrow$& PSNR / SSIM $\uparrow$& PSNR / SSIM $\uparrow$& PSNR / SSIM $\uparrow$& PSNR / SSIM $\uparrow$& PSNR / SSIM $\uparrow$\bigstrut\\
    \hline
    \multirow{4}[2]{*}{\makecell{Task\\Specific}} & QRNN3D \cite{wei20203}& 42.24 / 0.969 & 41.98 / 0.968 & 41.62 / 0.968 & 40.55 / 0.960 & 41.74 / 0.966 & 41.55 / 0.965 & 41.38 / 0.964 & 39.76 / 0.946 & 31.98 / 0.885 & 31.77 / 0.882 & 31.48 / 0.878 & 28.05 / 0.822 \bigstrut[t]\\
          & SST \cite{li2023spatial}  & 43.38 / 0.976 & 42.69 / 0.973 & 42.51 / 0.972 & 41.16 / 0.964 & 42.84 / 0.973 & 42.38 / 0.971 & 42.01 / \textcolor{blue}{0.970} & 40.56 / 0.956 & 33.85 / 0.907 & 33.69 / 0.905  & 33.37 / 0.901 & 29.92 / 0.841 \\
          & SERT \cite{li2023spectral} & \textcolor{blue}{43.96} / \textcolor{blue}{0.980} & \textcolor{blue}{43.48} / \textcolor{blue}{0.978} & \textcolor{blue}{43.45} / \textcolor{blue}{0.977} & \textcolor{blue}{42.37} / \textcolor{blue}{0.969} & \textcolor{blue}{43.56} / \textcolor{blue}{0.978} & \textcolor{blue}{43.19} / \textcolor{blue}{0.976} & \textcolor{blue}{42.88} / \textcolor{red}{0.974} & \textcolor{blue}{41.85} / \textcolor{blue}{0.963} & 34.48 / 0.922 & 34.26 / 0.920 & 33.98 / 0.915 & 30.53 / 0.854 \\
          & LDERT \cite{li2024latent}& \textcolor{red}{44.04} / \textcolor{red}{0.981} & \textcolor{red}{43.57} / \textcolor{red}{0.979} & \textcolor{red}{43.55} / \textcolor{red}{0.979} & \textcolor{red}{42.51} / \textcolor{red}{0.971}  & \textcolor{red}{43.67} / \textcolor{red}{0.979} & \textcolor{red}{43.33} / \textcolor{red}{0.977} & \textcolor{red}{43.06} / \textcolor{red}{0.974} & \textcolor{red}{42.02} / \textcolor{red}{0.965} & 34.65 / 0.923 & 34.42 / 0.920 & 34.13 / \textcolor{blue}{0.917} & 30.74 / 0.856 \bigstrut[b]\\
    \hline
    \multirow{7}[2]{*}{\makecell{All\\in\\One}} & AirNet \cite{li2022all}& 42.11 / 0.968 & 41.24 / 0.964 & 40.89 / 0.961 & 38.49 / 0.942 & 41.62 / 0.965 & 40.83 / 0.959 & 40.31 / 0.957 & 37.59 / 0.905 & 29.02 / 0.752 & 28.93 / 0.744 & 28.67 / 0.740 & 25.67 / 0.667 \bigstrut[t]\\
          & PromptIR \cite{potlapalli2023promptir}& 42.76 / 0.973 & 41.93 / 0.969 & 41.43 / 0.969 & 39.04 / 0.947 & 42.26 / 0.969 & 41.54 / 0.964 & 40.90 / 0.964 & 38.12 / 0.919 & 29.84 / 0.761 & 29.71 / 0.754 & 29.39 / 0.747 & 26.38 / 0.676 \\
          & PIP  \cite{li2023prompt} & 42.96 / 0.974 & 42.13 / 0.970 & 41.38 / 0.969 & 40.19 / 0.959 & 42.39 / 0.971 & 41.71 / 0.966 & 41.11 / 0.966 & 39.45 / 0.943 & 29.57 / 0.751 & 29.31 / 0.745 & 29.17 / 0.740 & 25.95 / 0.658 \\
          & HAIR \cite{cao2024hair} & 41.78 / 0.965 & 41.47 / 0.965 & 40.68 / 0.958 & 38.58 / 0.943 & 41.19 / 0.959 & 40.95 / 0.961 & 40.53 / 0.959 & 37.92 / 0.909 & 29.38 / 0.756 & 29.22 / 0.750 & 28.73 / 0.744 & 25.40 / 0.664 \\
          & InstructIR \cite{conde2024instructir}& 41.29 / 0.963 & 40.89 / 0.961 & 39.94 / 0.958 & 38.46 / 0.945 & 40.72 / 0.960 & 40.38 / 0.957 & 39.94 / 0.956 & 38.21 / 0.934 & 28.66 / 0.736 & 28.47 / 0.730 & 28.19 / 0.725 & 24.63 / 0.637 \\
          & PromptHSI \cite{lee2024prompthsi}& 40.61 / 0.967 & 40.36 / 0.965 & 39.30 / 0.960 & 36.27 / 0.927 & 40.22 / 0.951 & 39.98 / 0.955 & 39.42 / 0.953 & 35.39 / 0.884 & \textcolor{blue}{34.93} / \textcolor{blue}{0.931} & \textcolor{blue}{34.87} / \textcolor{blue}{0.930} & \textcolor{blue}{34.49} / \textcolor{red}{0.928} & \textcolor{blue}{30.78} / \textcolor{blue}{0.857} \\
          & MP-HSIR (Ours)  & 43.07 / 0.975 & 42.46 / 0.972 & 42.20 / 0.972 & 41.42 / 0.966 & 42.74  / 0.973 & 42.32 / 0.971 & 41.93 / \textcolor{blue}{0.970} & 40.98 / 0.960 & \textcolor{red}{35.21} / \textcolor{red}{0.933} & \textcolor{red}{34.99} / \textcolor{red}{0.931} & \textcolor{red}{34.72} / \textcolor{red}{0.928} & \textcolor{red}{31.36} / \textcolor{red}{0.880} \bigstrut[b]\\
    \hline
    \end{tabular}%
    }
   \caption{\textbf{[All-in-one]} Quantitative comparison of all-in-one and state-of-the-art task-specific methods under different cases on \textbf{\textit{Complex denoising}} tasks. The best and second-best performances are highlighted in \textcolor{red}{red} and \textcolor{blue}{blue}, respectively.}
  \label{tab:table6}%
\end{table*}%

\begin{figure}[t]
    \centering
    \includegraphics[width=0.48\textwidth]{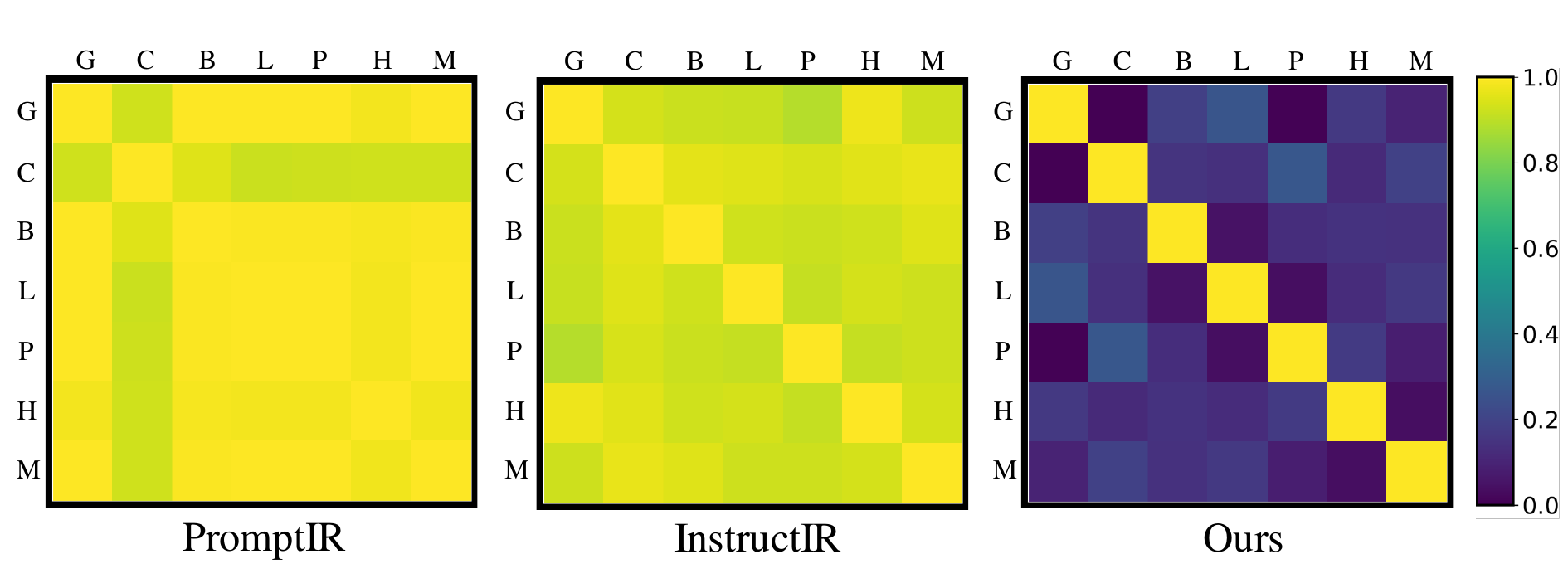} 
    \vskip -5pt
    \caption{Similarity matrices of prompt vector for seven HSI restoration tasks, with task names as in manuscript.}
    \label{fig:1}
\end{figure}

\section{Visualization Results of Prompts}

In Table \ref{fig:1}, we present the similarity matrices of prompt vectors for different degradation types, comparing PromptIR \cite{potlapalli2023promptir}, InstructIR \cite{conde2024instructir}, and the proposed method. PromptIR relies exclusively on visual prompts, InstructIR utilizes textual prompts for guidance, and the proposed method integrates text-visual synergistic prompts to enhance degradation modeling. As illustrated in the figure, both PromptIR and InstructIR struggle to effectively differentiate between various HSI degradation types, whereas our method demonstrates distinct separation for each individual task, highlighting its superior capability in handling diverse degradation scenarios.

\begin{table*}[!t]
  \centering
  \renewcommand{\arraystretch}{1.2} 
  \resizebox{\textwidth}{!}{
    \begin{tabular}{c|c|c|c|c|c|c|c|c|c|c}
    \hline
    \multirow{3}[6]{*}{\textbf{Type}} & \multirow{3}[6]{*}{\textbf{Methods}} & \multicolumn{3}{c|}{\textbf{\textit{Gaussian Deblurring} (ICVL \cite{arad2016sparse})}} & \multicolumn{3}{c|}{{\textbf{\textit{Gaussian Deblurring} (PaviaC \cite{huang2009comparative})}}} & \multicolumn{3}{c}{{\textbf{\textit{Gaussian Deblurring} (Eagle \cite{peerbhay2013commercial})}}} \bigstrut\\
\cline{3-11}          &       & Radius = 9 & Radius = 15 & Radius = 21 & Radius = 7 & Radius = 11 & Radius = 15 & Radius = 7 & Radius = 11 & Radius = 15 \bigstrut\\
\cline{3-11}          &       & PSNR / SSIM $\uparrow$& PSNR / SSIM $\uparrow$& PSNR / SSIM $\uparrow$& PSNR / SSIM $\uparrow$& PSNR / SSIM $\uparrow$& PSNR / SSIM $\uparrow$& PSNR / SSIM $\uparrow$& PSNR / SSIM $\uparrow$& PSNR / SSIM $\uparrow$\bigstrut\\
    \hline
    \multirow{4}[2]{*}{\makecell{Task\\Specific}} & Stripformer \cite{tsai2022stripformer}& 49.21 / 0.994 & 45.85 / 0.989 & 43.04 / 0.980 & 38.73 / 0.934 & 36.95 / 0.914 & 35.72 / 0.891 & 44.68 / 0.974 & 41.37 / 0.958 & 39.83 / 0.942 \bigstrut[t]\\
          & FFTformer \cite{kong2023efficient}& 49.83 / 0.994 & 46.43 / 0.989 & 43.69 / 0.981 & 39.82 / 0.942 & 37.63 / 0.919 & 36.44 / 0.901 & 45.51 / 0.980 & 42.18 / 0.962 & 40.59 / 0.944 \\
          & LoFormer \cite{mao2024loformer}& 50.35 / \textcolor{blue}{0.995} & 46.94 / \textcolor{blue}{0.991} & 44.15 / \textcolor{red}{0.983} & 39.54 / 0.941 & 37.41 / 0.918 & 36.21 / 0.892 & 45.32 / 0.979 & 42.03 / 0.961 & 40.42 / 0.946 \\
          & MLWNet \cite{gao2024efficient}& 50.68 / \textcolor{red}{0.996} & 47.54 / \textcolor{blue}{0.991} & \textcolor{blue}{44.76} / \textcolor{red}{0.983} & \textcolor{red}{40.85} / \textcolor{red}{0.949} & \textcolor{blue}{38.92} / \textcolor{red}{0.929} & \textcolor{red}{37.28} / 0.901 & \textcolor{red}{46.61} / \textcolor{red}{0.983} & \textcolor{red}{43.68} / \textcolor{red}{0.969} & \textcolor{red}{41.83} / \textcolor{red}{0.955} \bigstrut[b]\\
    \hline
    \multirow{7}[2]{*}{\makecell{All\\in\\One}} & AirNet \cite{li2022all}& 50.64 / \textcolor{blue}{0.995} & 47.10 / \textcolor{blue}{0.991} & 43.89 / \textcolor{blue}{0.982} & 39.45 / 0.940 & 37.76 / 0.921 & 36.08 / 0.893 & 44.83 / 0.977 & 42.32 / 0.966 & 40.11 / 0.946 \bigstrut[t]\\
          & PromptIR \cite{potlapalli2023promptir}& 51.00 / \textcolor{red}{0.996} & \textcolor{red}{47.70} / \textcolor{red}{0.992} & 44.32 / \textcolor{red}{0.983} & \textcolor{blue}{40.41} / \textcolor{blue}{0.948} & 38.62 / \textcolor{red}{0.929} & 37.13 / \textcolor{blue}{0.904} & 46.17 / 0.981 & 43.45 / \textcolor{blue}{0.968} & \textcolor{blue}{41.60} / \textcolor{red}{0.955} \\
          & PIP   \cite{li2023prompt}& \textcolor{blue}{51.05} / \textcolor{red}{0.996} & 47.29 / \textcolor{blue}{0.991} & 44.21 / \textcolor{red}{0.983} & 40.31 / 0.944 & 38.51 / 0.926 & 36.73 / 0.902 & 45.90 / \textcolor{blue}{0.982} & 42.75 / \textcolor{blue}{0.968} & 40.86 / 0.951 \\
          & HAIR  \cite{cao2024hair}& 49.43 / 0.994 & 46.06 / 0.989  & 43.87 / 0.981 & 39.97 / 0.943 & 38.18 / 0.923 & 35.64 / 0.891 & 45.14 / 0.979 & 42.69 / 0.966 & 40.48 / 0.947 \\
          & InstructIR \cite{conde2024instructir}& 24.19 / 0.533 & 33.44 / 0.878 & 44.51 / \textcolor{red}{0.983} & 19.66 / 0.329 & 25.12 / 0.583 & 36.32 / 0.894 & 23.35 / 0.477 & 34.05 / 0.872 & 40.64 / 0.949 \\
          & PromptHSI \cite{lee2024prompthsi}& 25.16 / 0.619 & 30.89 / 0.852 & 41.52 / 0.980 & 38.73 / 0.936 & 36.63 / 0.913 & 34.84 / 0.888 & 42.45 / 0.972 & 39.34 / 0.952 & 37.46 / 0.938 \\
          & MP-HSIR (Ours)  & \textcolor{red}{51.53} / \textcolor{red}{0.996} & \textcolor{blue}{47.60} / \textcolor{red}{0.992} & \textcolor{red}{45.07} / \textcolor{blue}{0.982} & \textcolor{red}{40.85} / \textcolor{red}{0.949} & \textcolor{red}{38.95} / \textcolor{blue}{0.928} & \textcolor{blue}{37.19} / \textcolor{red}{0.905} & \textcolor{blue}{46.26} / \textcolor{blue}{0.982} & \textcolor{blue}{43.54} / \textcolor{red}{0.969} & 41.36 / \textcolor{blue}{0.952} \bigstrut[b]\\
    \hline
    \end{tabular}%
    }
    \caption{\textbf{[All-in-one]} Quantitative comparison of all-in-one and state-of-the-art task-specific methods under different blur kernel radius on \textbf{\textit{Gaussian deblurring}} tasks. The best and second-best performances are highlighted in \textcolor{red}{red} and \textcolor{blue}{blue}, respectively.}
  \label{tab:table7}%
\end{table*}%

\section{More Experimental Results}

In this section, we present additional experimental results, including the results of more quantitative results, model efficiency, more results of ablation study, controllability analysis, and more visual comparisons.

\begin{table*}[!t]
  \centering
  \renewcommand{\arraystretch}{1.2} 
  \resizebox{\textwidth}{!}{
    \begin{tabular}{c|c|c|c|c|c|c|c|c|c|c}
    \hline
    \multirow{3}[6]{*}{\textbf{Type}} & \multirow{3}[6]{*}{\textbf{Methods}} & \multicolumn{3}{c|}{\textbf{\textit{Super-Resolution} (ARAD \cite{arad2022ntire})}} & \multicolumn{3}{c|}{\textbf{\textit{Super-Resolution} (PaviaU \cite{huang2009comparative})}} & \multicolumn{3}{c}{\textbf{\textit{Super-Resolution} (Houston \cite{wu2017convolutional})}} \bigstrut\\
\cline{3-11}          &       & Scale = 2 & Scale = 4 & Scale = 8 & Scale = 2 & Scale = 4 & Scale = 8 & Scale = 2 & Scale = 4 & Scale = 8 \bigstrut\\
\cline{3-11}          &       & PSNR / SSIM $\uparrow$& PSNR / SSIM $\uparrow$& PSNR / SSIM $\uparrow$& PSNR / SSIM $\uparrow$& PSNR / SSIM $\uparrow$& PSNR / SSIM $\uparrow$& PSNR / SSIM $\uparrow$& PSNR / SSIM $\uparrow$& PSNR / SSIM $\uparrow$\bigstrut\\
    \hline
    \multirow{4}[2]{*}{\makecell{Task\\Specific}} & SNLSR \cite{hu2024exploring}& 43.93 / 0.980 & 34.56 / 0.902 & 29.67 / 0.813 & 34.58 / 0.869 & 29.85 / 0.719 & 27.23 / 0.601 & 34.91 / 0.908 & 31.33 / 0.782 & 28.86 / 0.671 \bigstrut[t]\\
          & MAN \cite{wang2024multi}  & 44.81 / 0.985 & 35.35 / 0.912 & 30.49 / 0.830 & 34.92 / 0.872 & 30.26 / 0.723 & 27.59 / 0.604 & 35.26 / 0.911 & 31.68 / 0.785 & 29.15 / 0.677 \\
          & ESSAformer \cite{zhang2023essaformer}& 45.32 / 0.988 & 36.02 / 0.927 & 30.85 / 0.838 & 35.47 / 0.879 & 30.60 / 0.728 & 27.96 / 0.606 & 35.60 / 0.913 & 31.94 / 0.787 & 29.57 / 0.679 \\
          & SRFormer \cite{zhou2023srformer}& 45.84 / \textcolor{blue}{0.989} & 36.73 / \textcolor{blue}{0.931} & \textcolor{blue}{31.48} / \textcolor{blue}{0.845} & \textcolor{blue}{35.92} / \textcolor{blue}{0.887} & \textcolor{blue}{31.08} / \textcolor{blue}{0.745} & \textcolor{red}{28.41} / \textcolor{blue}{0.620} & \textcolor{blue}{36.15} / \textcolor{blue}{0.920} & 32.44 / 0.805 & 29.81 / 0.684 \bigstrut[b]\\
    \hline
    \multirow{7}[2]{*}{\makecell{All\\in\\One}} & AirNet \cite{li2022all}& 44.82 / 0.985 & 35.26 / 0.919 & 30.12 / 0.828 & 34.85 / 0.871 & 30.19 / 0.724 & 27.76 / 0.609 & 35.22 / 0.911 & 31.65 / 0.786 & 29.13 / 0.676 \bigstrut[t]\\
          & PromptIR \cite{potlapalli2023promptir}& 45.33 / 0.988 & 36.00 / 0.927 & 30.77 / 0.838 & 35.57 / 0.883 & 30.81 / 0.735 & 28.20 / 0.619 & 35.96 / 0.915 & 32.38 / 0.798 & 29.85 / 0.684 \\
          & PIP   \cite{li2023prompt}& \textcolor{blue}{46.01} / \textcolor{blue}{0.989} & \textcolor{red}{37.34} / \textcolor{red}{0.939} & \textcolor{red}{31.73} / \textcolor{red}{0.853} & 35.71 / 0.885 & 31.02 / 0.744 & 28.20 / 0.618 & 36.10 / 0.917 & \textcolor{blue}{32.55} / \textcolor{blue}{0.806} & \textcolor{red}{30.02} / \textcolor{red}{0.692} \\
          & HAIR  \cite{cao2024hair}& 43.77 / 0.984 & 35.89 / 0.924 & 30.87 / 0.836 & 35.49 / 0.882 & 30.79 / 0.736 & 28.13 / 0.616 & 35.81 / 0.913 & 32.17 / 0.795 & 29.64 / 0.680 \\
          & InstructIR \cite{conde2024instructir}& 43.47 / 0.984 & 35.46 / 0.921 & 30.61 / 0.834 & 35.25 / 0.879 & 30.71 / 0.732 & 28.05 / 0.613 & 35.68 / 0.909 & 32.15 / 0.789 & 29.73 / 0.681 \\
          & PromptHSI \cite{lee2024prompthsi}& 40.25 / 0.975 & 35.41 / \textcolor{blue}{0.931} & 29.35 / 0.806 & 34.84 / 0.871 & 30.13 / 0.722 & 27.27 / 0.602 & 35.34 / 0.912 & 31.62 / 0.778 & 28.59 / 0.635 \\
          & MP-HSIR (Ours)  & \textcolor{red}{46.72} / \textcolor{red}{0.991} & \textcolor{blue}{36.88} / \textcolor{red}{0.939} & 31.14 / 0.843 & \textcolor{red}{36.27} / \textcolor{red}{0.894} & \textcolor{red}{31.26} / \textcolor{red}{0.757} & \textcolor{blue}{28.38} / \textcolor{red}{0.630} & \textcolor{red}{36.57} / \textcolor{red}{0.926} & \textcolor{red}{32.68} / \textcolor{red}{0.813} & \textcolor{blue}{29.92} / \textcolor{blue}{0.690} \bigstrut[b]\\
    \hline
    \end{tabular}%
    }
  \caption{\textbf{[All-in-one]} Quantitative comparison of all-in-one and state-of-the-art task-specific methods under different downsampling scales on \textbf{\textit{Super-Resolution}} tasks. The best and second-best performances are highlighted in \textcolor{red}{red} and \textcolor{blue}{blue}, respectively.}
  \label{tab:table8}%
\end{table*}%

\begin{table}[htbp]
  \centering
  \renewcommand{\arraystretch}{0.9} 
  \tiny
  \resizebox{\columnwidth}{!}{
    \begin{tabular}{ccccc}
    \hline
    Dataset & PromptIR & InstructIR & PromptHSI & \textbf{MP-HSIR} \\
    \hline
    Urban & 14.95 & 15.56 & \underline{12.34} & \textbf{11.42} \\
    EO-1  & \underline{17.99} & 19.71 & 18.13 & \textbf{16.54} \\
    \hline
    \end{tabular}%
    }
  \caption{No-reference quality assessment on real datasets.}
  \label{tab:table18}%
\end{table}%
\begin{table*}[!t]
  \centering
  \renewcommand{\arraystretch}{1.0} 
  \resizebox{\textwidth}{!}{
    \begin{tabular}{c|c|c|c|c|c|c|c}
    \hline
    \multirow{3}[6]{*}{\textbf{Type}} & \multirow{3}[6]{*}{\textbf{Methods}} & \multicolumn{3}{c|}{\textbf{\textit{Inpainting} (ICVL \cite{arad2016sparse})}} & \multicolumn{3}{c}{\textbf{\textit{Inpainting} (Chikusei \cite{yokoya2016airborne})}} \bigstrut\\
\cline{3-8}          &       & Rate = 0.7 & Rate = 0.8 & Rate = 0.9 & Rate = 0.7 & Rate = 0.8 & Rate = 0.9 \bigstrut\\
\cline{3-8}          &       & PSNR / SSIM $\uparrow$& PSNR / SSIM $\uparrow$& PSNR / SSIM $\uparrow$& PSNR / SSIM $\uparrow$& PSNR / SSIM $\uparrow$& PSNR / SSIM $\uparrow$\bigstrut\\
    \hline
    \multirow{4}[2]{*}{\makecell{Task\\Specific}} & NAFNet \cite{chen2022simple}& 45.03 / 0.989 & 44.65 / 0.988 & 43.50 / 0.985 & \textcolor{blue}{40.34} / 0.952 & \textcolor{blue}{40.15} / \textcolor{blue}{0.955} & \textcolor{blue}{38.97} / \textcolor{blue}{0.952} \bigstrut[t]\\
          & Restormer \cite{zamir2022restormer}& 46.51 / 0.991 & 46.00 / 0.991 & 44.85 / \textcolor{blue}{0.988} & 36.52 / 0.902 & 36.34 / 0.892 & 36.13 / 0.903 \\
          & DDS2M \cite{miao2023dds2m}& 45.41 / 0.989 & 43.34 / 0.983 & 37.80 / 0.935 & 36.77 / 0.906 & 35.23 / 0.901 & 32.83 / 0.854 \\
          & HIR-Diff \cite{pang2024hir}& 41.82 / 0.973 & 38.87 / 0.949 & 36.04 / 0.924 & 38.59 / 0.923 & 37.96 / 0.920 & 36.41 / 0.904 \bigstrut[b]\\
    \hline
    \multirow{7}[2]{*}{\makecell{All\\in\\One}} & AirNet \cite{li2022all}& 43.10 / 0.983 & 43.06 / 0.983 & 41.65 / 0.977 & 38.12 / 0.919 & 37.86 / 0.921 & 36.39 / 0.918 \bigstrut[t]\\
          & PromptIR \cite{potlapalli2023promptir}& \textcolor{blue}{46.96} / \textcolor{blue}{0.992} & \textcolor{blue}{46.93} / \textcolor{blue}{0.992} & \textcolor{blue}{45.24} / \textcolor{blue}{0.988} & 38.86 / 0.925 & 38.30 / 0.931 & 37.05 / 0.934 \\
          & PIP   \cite{li2023prompt}& 44.37 / 0.985 & 43.47 / 0.983 & 42.26 / 0.978 & 38.74 / 0.922 & 38.58 / 0.930 & 37.98 / 0.938 \\
          & HAIR  \cite{cao2024hair}& 44.83 / 0.983 & 44.30 / 0.983 & 42.92 / 0.981 & 38.43 / 0.921 & 38.28 / 0.928 & 37.43 / 0.932 \\
          & InstructIR \cite{conde2024instructir}& 44.85 / 0.989 & 44.29 / 0.987 & 43.08 / 0.983 & 36.30 / 0.904 & 36.18 / 0.908 & 35.84 / 0.909 \\
          & PromptHSI \cite{lee2024prompthsi}& 42.83 / 0.983 & 41.72 / 0.976 & 39.89 / 0.956 & 38.99 / \textcolor{blue}{0.966} & 37.64 / 0.952 & 35.35 / 0.920 \\
          & MP-HSIR (Ours)  & \textcolor{red}{53.06} / \textcolor{red}{0.997} & \textcolor{red}{51.94} / \textcolor{red}{0.996} & \textcolor{red}{49.60} / \textcolor{red}{0.994} & \textcolor{red}{44.75} / \textcolor{red}{0.981} & \textcolor{red}{44.06} / \textcolor{red}{0.981} & \textcolor{red}{42.08} / \textcolor{red}{0.975} \bigstrut[b]\\
    \hline
    \end{tabular}%
    }
  \caption{\textbf{[All-in-one]} Quantitative comparison of all-in-one and state-of-the-art task-specific methods under different mask rates on \textbf{\textit{Inpainting}} tasks. The best and second-best performances are highlighted in \textcolor{red}{red} and \textcolor{blue}{blue}, respectively.}
  \label{tab:table9}%
\end{table*}%

\begin{table*}[!t]
  \centering
  \renewcommand{\arraystretch}{1.0} 
  \resizebox{\textwidth}{!}{
    \begin{tabular}{c|c|c|c|c|c|c|c}
    \hline
    \multirow{3}[6]{*}{\textbf{Type}} & \multirow{3}[6]{*}{\textbf{Methods}} & \multicolumn{3}{c|}{\textbf{\textit{Dehazing} (PaviaU \cite{huang2009comparative})}} & \multicolumn{3}{c}{\textbf{\textit{Dehazing} (Eagle \cite{peerbhay2013commercial})}} \bigstrut\\
\cline{3-8}          &       & Omega = 0.5 & Omega = 0.75 & Omega = 1.0 & Omega = 0.5 & Omega = 0.75 & Omega = 1.0 \bigstrut\\
\cline{3-8}          &       & PSNR / SSIM $\uparrow$& PSNR / SSIM $\uparrow$& PSNR / SSIM $\uparrow$& PSNR / SSIM $\uparrow$& PSNR / SSIM $\uparrow$& PSNR / SSIM $\uparrow$\bigstrut\\
    \hline
    \multirow{4}[2]{*}{\makecell{Task\\Specific}} & SGNet \cite{ma2022spectral}& 36.52 / 0.974 & 34.10 / 0.964 & 32.22 / 0.949 & 39.43 / 0.989 & 37.33 / 0.976 & 34.90 / 0.962 \bigstrut[t]\\
          & SCANet \cite{guo2023scanet}& 39.01 / 0.986 & 36.54 / 0.978 & 34.21 / 0.969 & 41.92 / 0.991 & 39.68 / 0.987 & 37.31 / 0.978 \\
          & MB-Taylor \cite{qiu2023mb}& \textcolor{blue}{40.51} / \textcolor{blue}{0.991} & \textcolor{blue}{38.03} / 0.984 & \textcolor{blue}{35.44} / 0.975 & \textcolor{blue}{43.76} / \textcolor{blue}{0.995} & \textcolor{blue}{40.97} / \textcolor{blue}{0.992} & 38.36 / 0.986 \\
          & DCMPNet \cite{zhang2024depth}& 39.63 / \textcolor{red}{0.993} & 37.14 / \textcolor{blue}{0.985} & 34.82 / \textcolor{blue}{0.976} & 42.93 / \textcolor{blue}{0.995} & 40.15 / 0.991 & 37.64 / 0.985 \bigstrut[b]\\
    \hline
    \multirow{7}[2]{*}{\makecell{All\\in\\One}} & AirNet \cite{li2022all}& 38.61 / 0.982 & 35.65 / 0.967 & 32.51 / 0.947 & 41.88 / 0.991 & 38.68 / 0.982 & 35.92 / 0.969 \bigstrut[t]\\
          & PromptIR \cite{potlapalli2023promptir}& 40.34 / \textcolor{blue}{0.991} & 37.43 / 0.983 & 34.47 / 0.971 & 43.55 / \textcolor{blue}{0.995} & 40.69 / \textcolor{blue}{0.992} & 37.94 / \textcolor{blue}{0.988} \\
          & PIP   \cite{li2023prompt}& 40.30 / \textcolor{blue}{0.991} & 37.64 / 0.983 & 34.93 / 0.971 & 43.21 / 0.994 & 40.95 / 0.991 & 38.07 / 0.985 \\
          & HAIR  \cite{cao2024hair}& 39.47 / 0.989 & 36.79 / 0.980 & 34.02 / 0.965 & 43.38 / \textcolor{blue}{0.995} & 40.54 / \textcolor{blue}{0.992} & \textcolor{blue}{38.67} / \textcolor{blue}{0.988} \\
          & InstructIR \cite{conde2024instructir}& 38.24 / 0.986 & 34.57 / 0.974 & 31.36 / 0.954 & 40.90 / 0.991 & 38.07 / 0.985 & 33.99 / 0.971 \\
          & PromptHSI \cite{lee2024prompthsi}& 38.62 / 0.982 & 36.48 / 0.975 & 35.22 / 0.964 & 40.88 / 0.986 & 40.49 / 0.984 & 37.98 / 0.981 \\
          & MP-HSIR (Ours)  & \textcolor{red}{42.64} / \textcolor{red}{0.993} & \textcolor{red}{39.46} / \textcolor{red}{0.988} & \textcolor{red}{36.68} / \textcolor{red}{0.978} & \textcolor{red}{45.66} / \textcolor{red}{0.997} & \textcolor{red}{42.24} / \textcolor{red}{0.995} & \textcolor{red}{39.34} / \textcolor{red}{0.992} \bigstrut[b]\\
    \hline
    \end{tabular}%
    }
  \caption{\textbf{[All-in-one]} Quantitative comparison of all-in-one and state-of-the-art task-specific methods under different haze levels on \textbf{\textit{Dehazing}} tasks. The best and second-best performances are highlighted in \textcolor{red}{red} and \textcolor{blue}{blue}, respectively.}
  \label{tab:table10}%
\end{table*}%

\begin{table*}[!t]
  \centering
  \renewcommand{\arraystretch}{1.0} 
  \resizebox{\textwidth}{!}{
    \begin{tabular}{c|c|c|c|c|c|c|c}
    \hline
    \multirow{3}[6]{*}{\textbf{Type}} & \multirow{3}[6]{*}{\textbf{Methods}} & \multicolumn{3}{c|}{\textbf{\textit{Band Completion} (ARAD \cite{arad2022ntire})}} & \multicolumn{3}{c}{\textbf{\textit{Band Completion} (Berlin \cite{okujeni2016berlin})}} \bigstrut\\
\cline{3-8}          &       & Rate = 0.1 & Rate = 0.2 & Rate = 0.3 & Rate = 0.1 & Rate = 0.2 & Rate = 0.3 \bigstrut\\
\cline{3-8}          &       & PSNR / SSIM $\uparrow$& PSNR / SSIM $\uparrow$& PSNR / SSIM $\uparrow$& PSNR / SSIM $\uparrow$& PSNR / SSIM $\uparrow$& PSNR / SSIM $\uparrow$\bigstrut\\
    \hline
    \multirow{4}[2]{*}{\makecell{Task\\Specific}} & NAFNet \cite{chen2022simple}& 47.82 / 0.996 & 47.02 / 0.995 & 46.29 / 0.994 & 39.71 / 0.972 & 37.51 / 0.874 & 37.84 / 0.875 \bigstrut[t]\\
          & Restormer \cite{zamir2022restormer}& 49.24 / 0.997 & 48.29 / 0.995 & 47.49 / 0.993 & 34.98 / 0.605 & 35.24 / 0.606 & 35.00 / 0.607 \\
          & SwinIR \cite{liang2021swinir}& 50.95 / 0.997 & 49.80 / 0.995 & 48.49 / 0.993 & 36.88 / 0.950 & 34.69 / 0.855 & 34.78 / 0.854 \\
          & MambaIR \cite{guo2024mambair}& 51.46 / \textcolor{blue}{0.998} & 50.32 / 0.995 & 49.01 / 0.993 & 37.54 / 0.953 & 35.38 / 0.857 & 35.43 / 0.855 \bigstrut[b]\\
    \hline
    \multirow{7}[2]{*}{\makecell{All\\in\\One}} & AirNet \cite{li2022all}& 46.14 / 0.994 & 45.21 / 0.992 & 44.46 / 0.990 & 37.62 / 0.691 & 35.26 / 0.586 & 34.86 / 0.595 \bigstrut[t]\\
          & PromptIR \cite{potlapalli2023promptir}& 47.68 / 0.996 & 46.71 / 0.994 & 45.41 / 0.992 & 42.81 / 0.707 & 39.55 / 0.910 & 39.00 / 0.640 \\
          & PIP  \cite{li2023prompt} & 48.35 / 0.995 & 47.37 / 0.994 & 46.37 / 0.991 & 38.60 / 0.706 & 36.58 / 0.657 & 36.43 / 0.641 \\
          & HAIR \cite{cao2024hair} & 46.27 / 0.994 & 44.92 / 0.992 & 44.04 / 0.990 & 40.04 / 0.705 & 36.54 / 0.607 & 37.45 / 0.639 \\
          & InstructIR \cite{conde2024instructir}& \textcolor{blue}{52.66} / \textcolor{blue}{0.998} & \textcolor{blue}{51.37} / \textcolor{blue}{0.997} & \textcolor{blue}{49.90} / \textcolor{blue}{0.996} & 36.17 / 0.606 & 35.33 / 0.559 & 36.40 / 0.576 \\
          & PromptHSI \cite{lee2024prompthsi}& 49.05 / 0.996 & 47.09 / 0.993 & 45.89 / 0.992 & \textcolor{blue}{47.11} / \textcolor{blue}{0.997} & \textcolor{blue}{43.14} / \textcolor{blue}{0.973} & \textcolor{blue}{39.82} / \textcolor{blue}{0.956} \\
          & MP-HSIR (Ours)  & \textcolor{red}{57.83} / \textcolor{red}{0.999} & \textcolor{red}{56.61} / \textcolor{red}{0.999} & \textcolor{red}{54.99} / \textcolor{red}{0.998} & \textcolor{red}{52.14} / \textcolor{red}{0.999} & \textcolor{red}{49.20} / \textcolor{red}{0.997} & \textcolor{red}{47.26} / \textcolor{red}{0.965} \bigstrut[b]\\
    \hline
    \end{tabular}%
    }
  \caption{\textbf{[All-in-one]} Quantitative comparison of all-in-one and state-of-the-art task-specific methods under different mask rates on \textbf{\textit{Band Completion}} tasks. The best and second-best performances are highlighted in \textcolor{red}{red} and \textcolor{blue}{blue}, respectively.}
  \label{tab:table11}%
\end{table*}%

\subsection{More Quantitative Results}
In this section, we provide a detailed quantitative comparison for 7 all-in-one HSI restoration tasks. The experimental results across different degradation levels are systematically presented for each task, including Gaussian denoising in Table \ref{tab:table5}, complex denoising in Table \ref{tab:table6}, Gaussian deblurring in Table \ref{tab:table7}, super-resolution in Table \ref{tab:table8}, inpainting in Table \ref{tab:table9}, dehazing in Table \ref{tab:table10}, and band completion in Table \ref{tab:table11}.

\begin{figure}[t]
    \centering
    \includegraphics[width=0.48\textwidth]{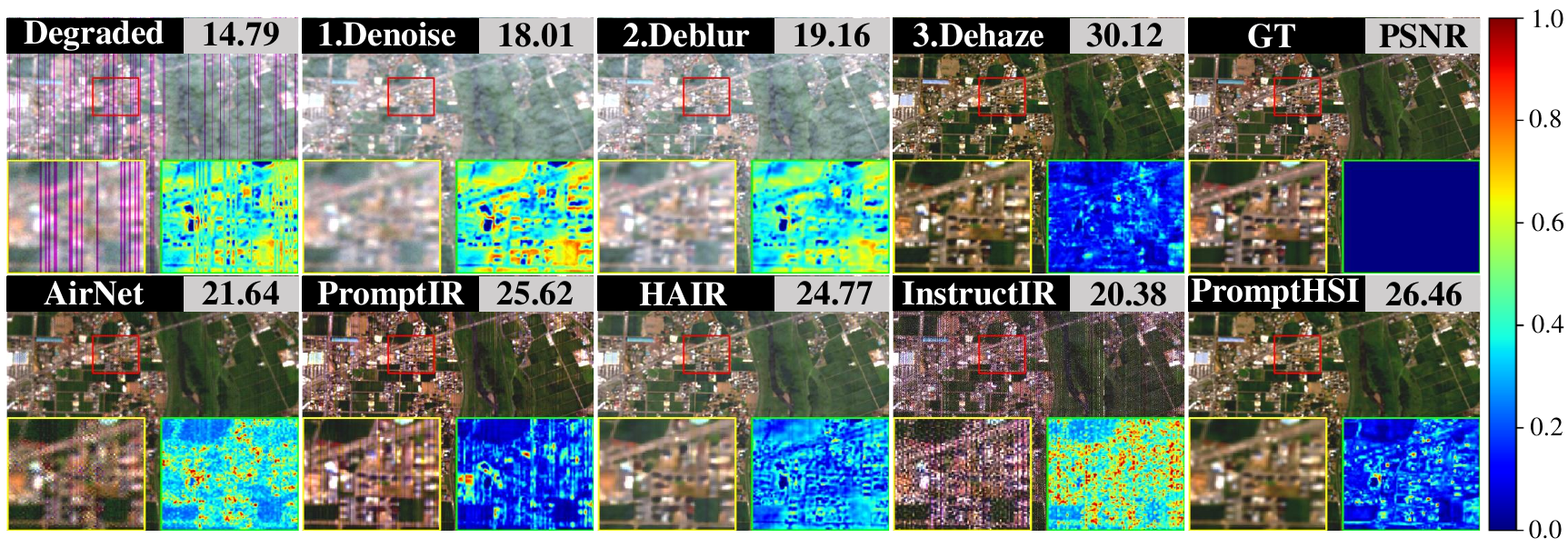} 
    \caption{Progressive results (ours) and final results (others).}
    \label{fig:4}
\end{figure}

In addition, we further evaluate the performance on real-world hyperspectral data using the no-reference metric QSFL \cite{yang2017no}. The results on the Urban and EO-1 datasets are reported in Table~\ref{tab:table18}, providing a more comprehensive assessment of generalization ability under practical conditions.

\begin{table}[!t]
  \centering
  \renewcommand{\arraystretch}{1.2} 
  \resizebox{\columnwidth}{!}{
    \begin{tabular}{ccccc}
    \hline
    \multirow{2}[4]{*}{\textbf{Methods}} & \multicolumn{2}{c}{\textbf{Natural Scene}} & \multicolumn{2}{c}{\textbf{Remote Sensing}} \bigstrut\\
\cline{2-5}          & Params (M) & FLOPS (G) & Params (M) & FLOPS (G) \bigstrut\\
    \hline
    AirNet \cite{li2022all}& 5.82  & 19.04  & 12.23  & 43.79  \bigstrut[t]\\
    PromptIR \cite{potlapalli2023promptir}& 33.00  & 10.03  & 72.60  & 22.21  \\
    PIP   \cite{li2023prompt}& 27.80  & 10.66  & 58.26  & 22.08  \\
    HAIR \cite{cao2024hair} & 7.68  & 2.72  & 17.28  & 6.46  \\
    InstructIR \cite{conde2024instructir}& 68.82  & 2.81  & 154.03  & 6.57  \\
    PromptHSI \cite{lee2024prompthsi}& 25.90  & 10.10  & 50.89  & 21.91  \\
    \rowcolor{gray!20} 
    MP-HSIR (Ours) & 13.88  & 14.40  & 30.91  & 32.74  \bigstrut[b]\\
    \hline
    \end{tabular}%
    }
  \caption{Model complexity comparisons}
  \label{tab:table12}%
\end{table}%

\begin{table}[htbp]
  \centering
  \renewcommand{\arraystretch}{1.0} 
  \resizebox{\columnwidth}{!}{
    \begin{tabular}{ccccc}
    \hline
    Metric & PromptIR & InstructIR & PromptHSI & \textbf{MP-HSIR} \\
    \hline
    Inference Time (s) & 0.087  & \textbf{0.065}  & 0.282  & \underline{0.083}  \\
    \hline
    \end{tabular}%
    }
  \caption{Inference time for remote sensing scenes (64×64×100).}
  \label{tab:table17}%
\end{table}%

\begin{table}[htbp]
  \centering
  \renewcommand{\arraystretch}{1.3} 
  \resizebox{\columnwidth}{!}{
    \begin{tabular}{lllccc}
    \hline
    \multicolumn{3}{l}{\textbf{Method}} & PSNR $\uparrow$  & SSIM $\uparrow$  & Params (M) \\
    \hline
    \multicolumn{3}{l}{Baseline (Only Spatial SA)} & 33.78  & 0.782  & 20.93  \\
    \hline
    \multicolumn{3}{l}{+ Textual Prompt $P_T$} & 34.53  & 0.807  & 21.51  \\
    \multicolumn{3}{l}{+ Visual Prompt $P_V$} & 34.47  & 0.805  & 23.68  \\
    \multicolumn{3}{l}{+ Textual Prompt $P_T$ + Visual Prompt $P_V$} & 34.92  & 0.822  & 24.26  \\
    \hline
    \multicolumn{3}{l}{+ Global Spectral SA + $P_T$ + $P_T$} & 35.20  & 0.835  & 30.07  \\
    \multicolumn{3}{l}{+ Local Spectral SA + $P_T$ + $P_V$} & 35.82  & 0.846  & 24.43  \\
    \multicolumn{3}{l}{+ Local Spectral SA + $P_T$ + $P_V$ + Spectral Prompt $P_S$} & 36.67  & 0.863  & 25.10  \\
    \hline
    \multicolumn{3}{l}{Full Model} & \textbf{37.17}  & \textbf{0.874}  & 30.91  \\
    \hline
    \end{tabular}%
    }
  \caption{Ablation study to verify the effectiveness of modules on Xiong’an dataset in \textbf{\textit{Gaussian denoising}} task with sigma = 70.}
  \label{tab:table14}%
\end{table}%

\subsection{Model Efficiency}
In this section, we present the parameter counts and computational costs of the all-in-one models for both natural scene and remote sensing hyperspectral datasets. Notably, the network width for remote sensing datasets is 1.5 times greater than that for natural scene datasets across all models. As demonstrated in Table \ref{tab:table12}, our method achieves a lower parameter count while maintaining competitive computational efficiency.

Furthermore, to evaluate the practical inference performance, Table~\ref{tab:table17} reports the average inference time of each method on remote sensing datasets. Our method maintains a favorable trade-off between efficiency and accuracy, demonstrating its suitability for large-scale deployment.

\subsection{More results of Ablation Study}
In this section, we present ablation studies on the textual prompts $P_T$, learnable visual prompts $P_V$, global spectral self-attention, local spectral self-attention, and spectral prompts $P_S$ across multiple tasks, as shown in Tables \ref{tab:table14}, \ref{tab:table15}, and \ref{tab:table16}. Overall, the addition of each module progressively improves the two accuracy metrics across all tasks.

\begin{table}[!t]
  \centering
  \renewcommand{\arraystretch}{1.3} 
  \resizebox{\columnwidth}{!}{
    \begin{tabular}{lllccc}
    \hline
    \multicolumn{3}{l}{\textbf{Method}} & PSNR $\uparrow$  & SSIM $\uparrow$  & Params (M) \\
    \hline
    \multicolumn{3}{l}{Baseline (Only Spatial SA)} & 30.10  & 0.788  & 20.93  \\
    \hline
    \multicolumn{3}{l}{+ Textual Prompt $P_T$} & 37.64  & 0.928  & 21.51  \\
    \multicolumn{3}{l}{+ Visual Prompt $P_V$} & 37.42  & 0.925  & 23.68  \\
    \multicolumn{3}{l}{+ Textual Prompt $P_T$ + Visual Prompt $P_V$} & 38.91  & 0.942  & 24.26  \\
    \hline
    \multicolumn{3}{l}{+ Global Spectral SA + $P_T$ + $P_T$} & 39.20  & 0.944  & 30.07  \\
    \multicolumn{3}{l}{+ Local Spectral SA + $P_T$ + $P_V$} & 39.79  & 0.948  & 24.43  \\
    \multicolumn{3}{l}{+ Local Spectral SA + $P_T$ + $P_V$ + Spectral Prompt $P_S$} & 40.51 & 0.950  & 25.10  \\
    \hline
    \multicolumn{3}{l}{Full Model} & \textbf{41.36}  & \textbf{0.952}  & 30.91  \\
    \hline
    \end{tabular}%
    }
  \caption{Ablation study to verify the effectiveness of modules on Eagle dataset in \textbf{\textit{Gaussian deblurring}} task with radius = 15.}
  \label{tab:table15}%
\end{table}%

\begin{table}[!t]
  \centering
  \renewcommand{\arraystretch}{1.3} 
  \resizebox{\columnwidth}{!}{
    \begin{tabular}{lllccc}
    \hline
    \multicolumn{3}{l}{\textbf{Method}} & PSNR $\uparrow$  & SSIM $\uparrow$  & Params (M) \\
    \hline
    \multicolumn{3}{l}{Baseline (Only Spatial SA)} & 32.52  & 0.967  & 20.93  \\
    \hline
    \multicolumn{3}{l}{+ Textual Prompt $P_T$} & 34.24  & 0.965  & 21.51  \\
    \multicolumn{3}{l}{+ Visual Prompt $P_V$} & 34.13  & 0.964  & 23.68  \\
    \multicolumn{3}{l}{+ Textual Prompt $P_T$ + Visual Prompt $P_V$} & 34.92  & 0.969  & 24.26  \\
    \hline
    \multicolumn{3}{l}{+ Global Spectral SA + $P_T$ + $P_T$} & 35.53  & 0.973  & 30.07  \\
    \multicolumn{3}{l}{+ Local Spectral SA + $P_T$ + $P_V$} & 35.64  & 0.974  & 24.43  \\
    \multicolumn{3}{l}{+ Local Spectral SA + $P_T$ + $P_V$ + Spectral Prompt $P_S$} & 36.13 & 0.976  & 25.10  \\
    \hline
    \multicolumn{3}{l}{Full Model} & \textbf{36.68}  & \textbf{0.978}  & 30.91  \\
    \hline
    \end{tabular}%
    }
  \caption{Ablation study to verify the effectiveness of modules on PaviaU dataset in \textbf{\textit{Dehazing}} task with Omega = 1.0.}
  \label{tab:table16}%
\end{table}%

\begin{figure}[tp]
    \centering
    \includegraphics[width=0.48\textwidth]{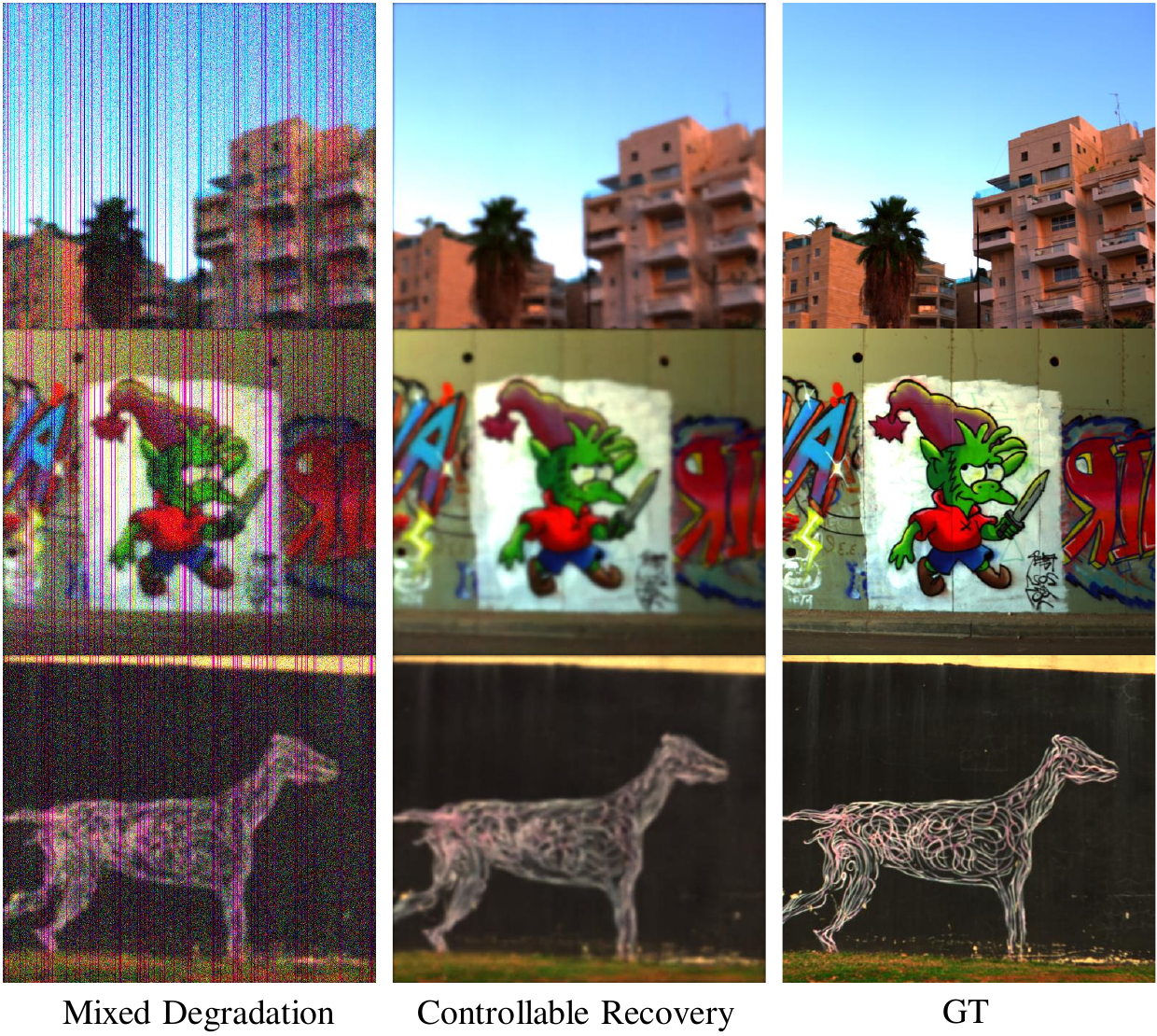} 
    \caption{Controllable Reconstruction: Removing \textbf{\textit{Gaussian }} from \textbf{\textit{Gaussian Noise}} + \textbf{\textit{Gaussian Blur}} Degradation.}
    \label{fig:2}
\end{figure}

\begin{figure}[tp]
    \centering
    \includegraphics[width=0.48\textwidth]{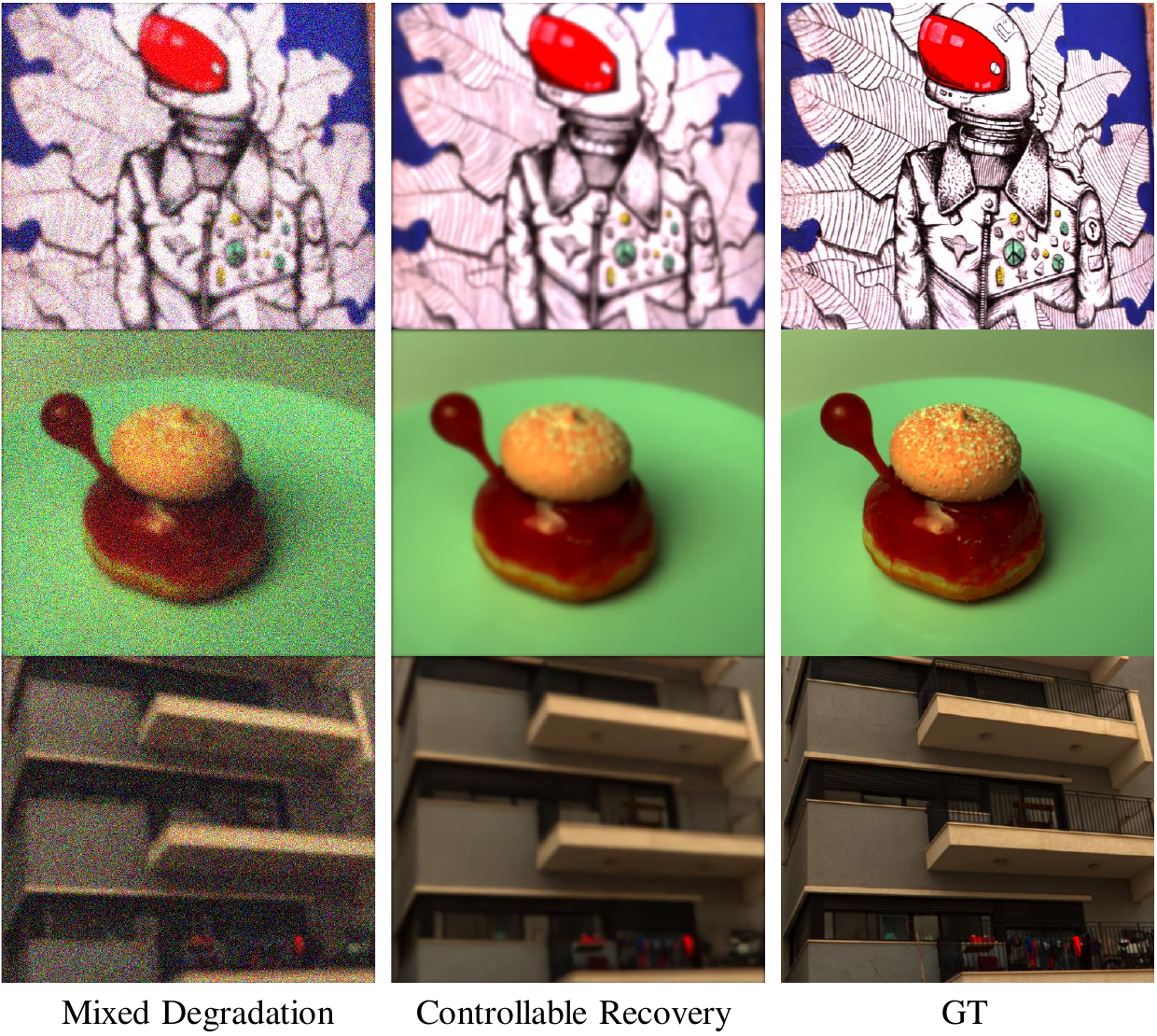} 
    \caption{Controllable Reconstruction: Removing \textbf{\textit{Complex Noise}} from \textbf{\textit{Complex Noise}} + \textbf{\textit{Gaussian Blur}} Degradation.}
    \label{fig:3}
\end{figure}

\subsection{Controllable Results}
In this section, we demonstrate the controllable restoration capability of the proposed method. Specifically, we conduct controlled restoration tasks under two composite degradation scenarios: Gaussian noise with Gaussian blur and complex noise with Gaussian blur, aiming to remove noise while preserving blur. As illustrated in Figures \ref{fig:2} and \ref{fig:3}, our method can precisely remove specific degradation types through accurate guidance from textual prompts, highlighting its controllability and interpretability.

Building on this controllable restoration paradigm, we further explore sequential degradation removal, where multiple degradations are addressed step by step under prompt guidance. As shown in Figure~\ref{fig:4}, our method achieves superior results in this more challenging setting, outperforming other approaches in both visual quality and flexibility.

\begin{figure*}[tp]
    \centering
    \includegraphics[width=1.0\textwidth]{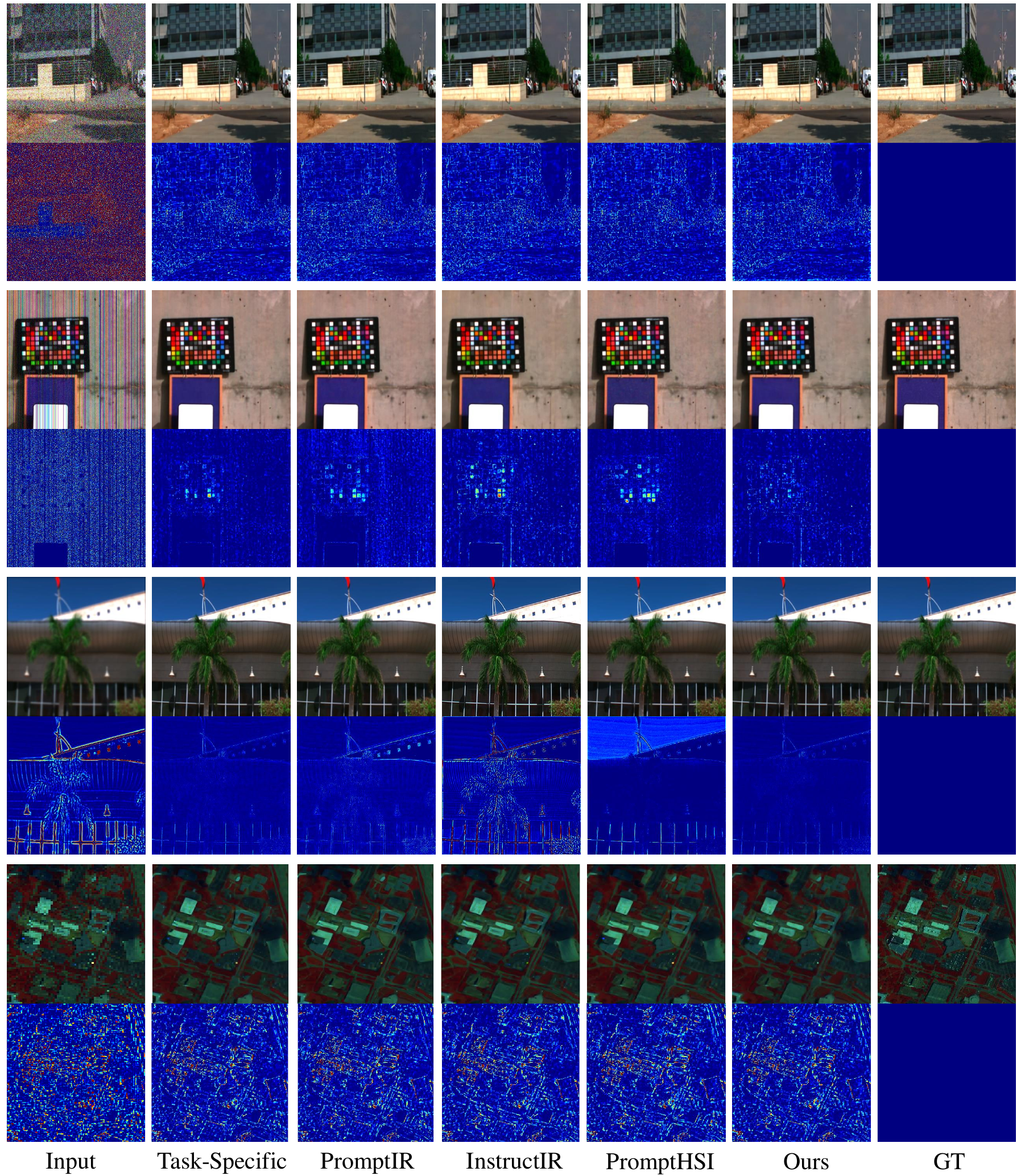}
    \vspace{3pt}
    \caption{Visual comparison results of \textbf{\textit{Gaussian denoising}}, \textbf{\textit{complex denoising}}, \textbf{\textit{Gaussian deblurring}}, and \textbf{\textit{super-resolution}}, including the corresponding residual maps. Task-Specific represents the optimal task-specific method.}
    \label{fig:5}
\end{figure*}

\begin{figure*}[tp]
    \centering
    \includegraphics[width=1.0\textwidth]{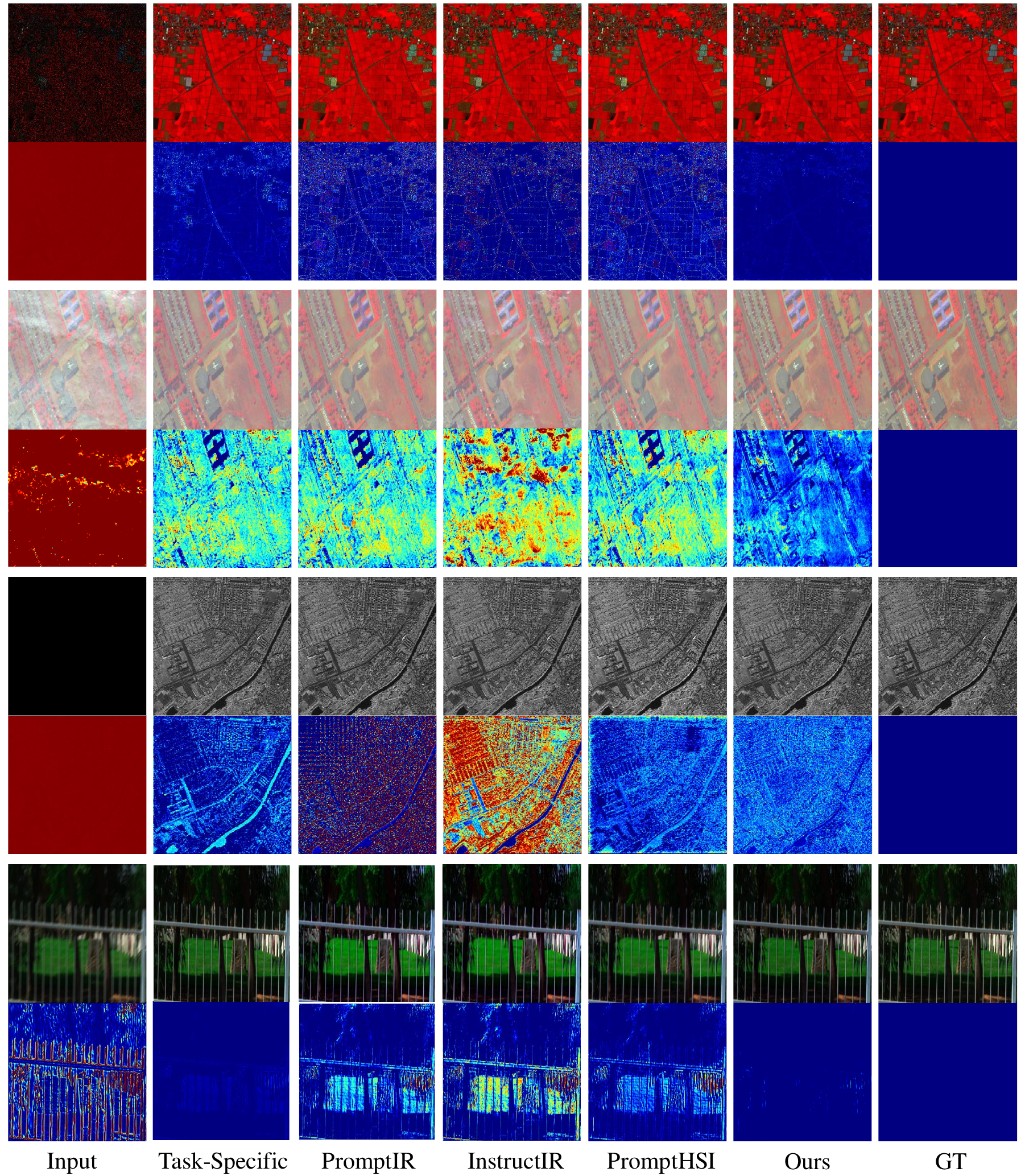}
    \vspace{3pt}
    \caption{Visual comparison results of \textbf{\textit{Inpainting}}, \textbf{\textit{Dehazing}}, \textbf{\textit{Band Completion}}, and \textbf{\textit{Motion Deblurring}}, including the corresponding residual maps. Task-Specific represents the optimal task-specific method.}
    \label{fig:6}
\end{figure*}

\begin{figure*}[tp]
    \centering
    \includegraphics[width=1.0\textwidth]{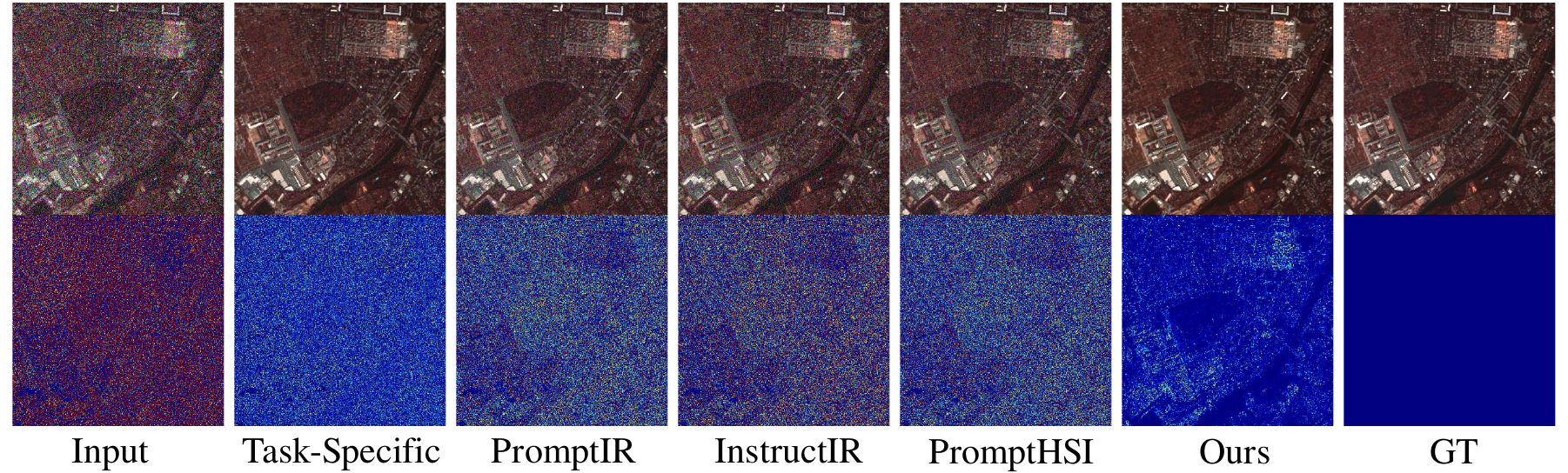} 
    \caption{Visual comparison results of \textbf{\textit{Poisson Denoising}}, including the corresponding residual maps. Task-Specific represents the optimal task-specific method.}
    \label{fig:7}
\end{figure*}

\begin{figure*}[tp]
    \centering
    \includegraphics[width=1.0\textwidth]{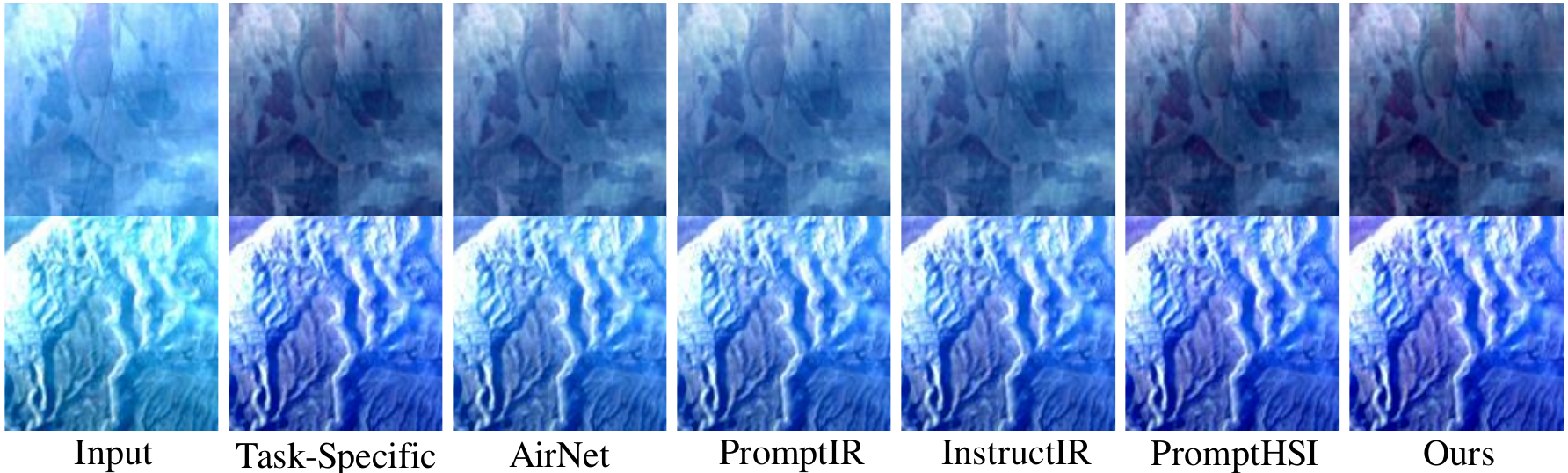} 
    \caption{Visual comparison results of  \textbf{\textit{Real Dehazing}}. Task-Specific represents the optimal task-specific method.}
    \label{fig:8}
\end{figure*}

\subsection{More Visual Results}
In this section, we present further visual results for each task, including all-in-one experiments, generalization testing, and real-world scenarios. As shown in Figures \ref{fig:5},  \ref{fig:6}, \ref{fig:7}, and \ref{fig:8}, the visualization results indicate that our method achieves the best performance in restoring texture details and structural features.  

\end{document}